\documentclass[lettersize,journal]{IEEEtran}
\usepackage{amsmath,amsfonts}
\usepackage{algorithmic}
\usepackage{algorithm}
\usepackage{array}
\usepackage[caption=false,font=normalsize,labelfont=sf,textfont=sf]{subfig}
\usepackage{textcomp}
\usepackage{stfloats}
\usepackage{url}
\usepackage{verbatim}
\usepackage{graphicx}
\usepackage{cite}
\usepackage{tabu}                      
\usepackage{booktabs}                  
\usepackage{lipsum}                    
\usepackage{mwe}                       
\usepackage{tabularray}

\usepackage{mathptmx}                  
\usepackage{amsmath} 
\usepackage{amssymb}  
\usepackage{amsfonts} 
\usepackage{mathptmx}
\usepackage{soul}
\usepackage{color, xcolor}
\soulregister{\cite}7
\soulregister{\citep}7
\soulregister{\citet}7
\soulregister{\ref}7
\soulregister{\pageref}7
\DeclareMathAlphabet{\mathcal}{OMS}{cmsy}{m}{n}
\DeclareSymbolFont{largesymbols}{OMX}{cmex}{m}{n}

\definecolor{c1st}{RGB}{255, 128, 128}
\definecolor{c2nd}{RGB}{255, 191, 128}
\definecolor{c3rd}{RGB}{245, 245, 0}

\graphicspath{{figures/}{pictures/}{images/}{./}} 

\begin{document}

\title{GeoTexDensifier: Geometry-Texture-Aware Densification for High-Quality Photorealistic 3D Gaussian Splatting}

\author{Hanqing Jiang, Xiaojun Xiang, Han Sun, Hongjie Li, Liyang Zhou, Xiaoyu Zhang,\\Guofeng Zhang, \textit{Member, IEEE}

\thanks{Hanqing Jiang, Xiaojun Xiang, Han Sun, Hongjie Li, Liyang Zhou and Xiaoyu Zhang are with SenseTime Research. E-mails: \{jianghanqing,xiangxiaojun,sunhan,lihongjie2,zhouliyang,zhangxiaoyu\}\\@sensetime.com.}
\thanks{Guofeng Zhang is with the State Key Lab of CAD\&CG, Zhejiang University. E-mail: zhangguofeng@zju.edu.cn.}
\thanks{Corresponding Author: Guofeng Zhang.}
\thanks{Hanqing Jiang, Xiaojun Xiang and Han Sun assert equal contribution and joint first authorship.}}

\markboth{IEEE Transactions on Visualization and Computer Graphics,~Vol.~XX, No.~XX, December~2024}%
{Hanqing Jiang \MakeLowercase{\textit{et al.}}: GeoTexDensifier: Geometry-Texture-Aware Densification for High-Quality Photorealistic 3D Gaussian Splatting}

\IEEEpubid{
\begin{minipage}{\textwidth}
\centering
Copyright~\copyright~2024 IEEE. Personal use of this material is permitted. \\However, permission to use this material for any other purposes must be obtained from the IEEE by sending an email to pubs-permissions@ieee.org.
\end{minipage}
}

\maketitle

\begin{abstract}
3D Gaussian Splatting (3DGS) has recently attracted wide attentions in various areas such as 3D navigation, Virtual Reality (VR) and 3D simulation, due to its photorealistic and efficient rendering performance. High-quality reconstrution of 3DGS relies on sufficient splats and a reasonable distribution of these splats to fit real geometric surface and texture details, which turns out to be a challenging problem. We present GeoTexDensifier, a novel geometry-texture-aware densification strategy to reconstruct high-quality Gaussian splats which better comply with the geometric structure and texture richness of the scene. Specifically, our GeoTexDensifier framework carries out an auxiliary texture-aware densification method to produce a denser distribution of splats in fully textured areas, while keeping sparsity in low-texture regions to maintain the quality of Gaussian point cloud. Meanwhile, a geometry-aware splitting strategy takes depth and normal priors to guide the splitting sampling and filter out the noisy splats whose initial positions are far from the actual geometric surfaces they aim to fit, under a \textit{Validation of Depth Ratio Change} checking. With the help of relative monocular depth prior, such geometry-aware validation can effectively reduce the influence of scattered Gaussians to the final rendering quality, especially in regions with weak textures or without sufficient training views. The texture-aware densification and geometry-aware splitting strategies are fully combined to obtain a set of high-quality Gaussian splats. We experiment our GeoTexDensifier framework on various datasets and compare our Novel View Synthesis results to other state-of-the-art 3DGS approaches, with detailed quantitative and qualitative evaluations to demonstrate the effectiveness of our method in producing more photorealistic 3DGS models. 
\end{abstract}

\begin{IEEEkeywords}
3D Gaussian Splatting, adaptive density control, texture-aware densification, geometry-aware splitting.
\end{IEEEkeywords}

\section{Introduction}
\IEEEPARstart{R}{ecent} years, 3D Gaussian Splatting (3DGS)~\cite{kerbl20233d} has attracted widespread attentions due to its efficient rendering performance and photorealistic visualization effects, and shown potential usefulness in various areas such as 3D navigation, Virtual Reality (VR), Mixed Reality (MR), 3D simulation and digital twins. Compared to the traditional Multi-View Stereo (MVS)~\cite{Seitz:2006:ACE, schoenberger2016mvs} and texture mapping~\cite{fuhrmann2014mve} methods which reconstruct explicit mesh models, 3DGS models are able to present more realistic texture and appearance details in online performance. Additionally, compared to the implicit representation of Neural Radiance Fields (NeRF) like \cite{mildenhall2021nerf,barron2022mip,barron2022mip,barron2023zip}, 3DGS innovatively proposes to use a set of differentiable 3D Gaussian ellipsoids called splats to represent the explicit structure of the captured scene, which more friendly support graphics techniques like editing~\cite{guedon2024sugar,waczynska2024games,gao2024mani,wang2024gaussianeditor}, relighting~\cite{gao2023relightable} and physical simulation~\cite{xie2024physgaussian} of the Gaussian splats in 3D space.

\IEEEpubidadjcol

3DGS achieves complete scene reconstruction and texture detail enhancement through the splitting and cloning of initial Gaussian points. High-quality reconstruction and rendering of 3DGS models rely on two conditions: the first is to have a sufficient number of Gaussian splats to support appearance details,
and the second is to ensure that the splats are optimized to the correct positions in 3D space. Currently, very few studies have explored the improvement of Gaussian splat densification strategy. For instance, Mini-Splatting~\cite{fang2024mini} introduces blur split strategy to further densify Gaussians with very large scales, but might generate a denser and more uniform spatial distribution with too many Gaussians, which is alleviated by additional simplification. An optimal but challenging densification strategy is to further densify splat distribution in fully textured areas, and still leave Gaussians sparse in more weakly textured regions. There are even fewer works discussing the geometrically accurate positioning of Gaussian splats. GeoGaussian~\cite{li2024geogaussian} transfers thin splats aligned with the smoothly connected areas observed from point cloud to newly generated ones through a carefully designed densification strategy, but relies heavily on the surfaces detected from the recovered splats in textureless regions which are usually noisy. Actually, for textured areas, there are usually sufficient multi-view visual cues for optimizing the Gaussian points to the accurate positions or remove the incorrect ones during the training process, while in textureless regions however, the lack of adequate visual observations makes it challenging to ensure the correct positioning of Gaussian splats.

\begin{figure*}[tbh]
  \centering
  \includegraphics[width=\linewidth]{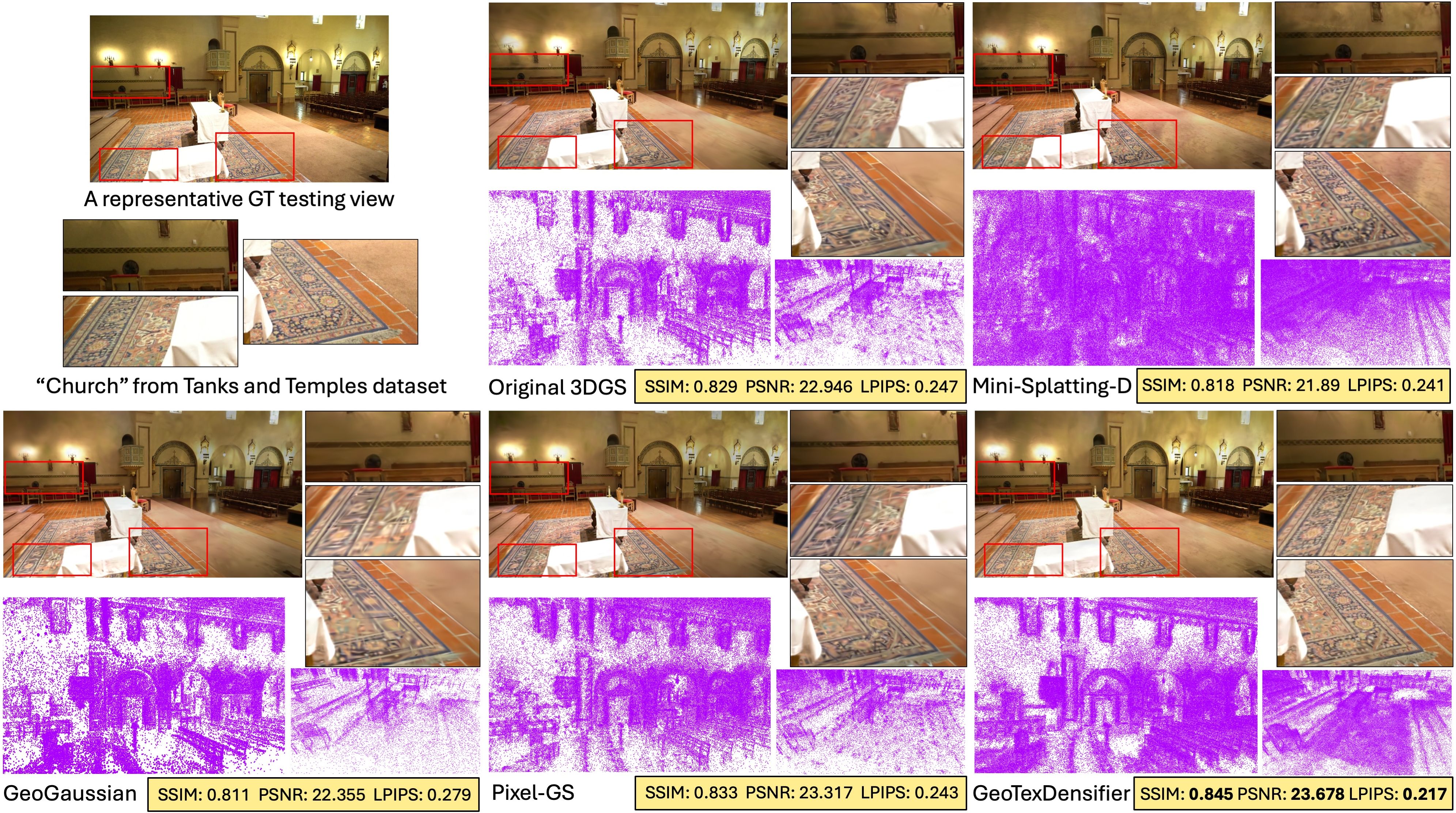}
  \caption{3DGS reconstruction of the case ``Church'' from Tanks and Temples dataset~\cite{knapitsch2017tanks}. A representative source image is taken as GT testing view. The reconstructed 3D Gaussian point clouds by original 3DGS~\cite{kerbl20233d}, Mini-Splatting-D~\cite{fang2024mini}, GeoGaussian~\cite{li2024geogaussian}, Pixel-GS~\cite{zhang2024pixel} and our GeoTexDensifier are given together with their respective rendered images in the testing views, to show that our framework performs the best in both spatial distribution of Gaussian splats and photorealistic rendering results, as verified by the rendering quality evaluation on SSIM, PSNR in dB, and LPIPS.}
  \label{fig:teaser}
\end{figure*}

To better tackle both densification and positioning problems mentioned above, this paper presents a novel 3D Gaussian splatting framework for high-quality photorealistic novel view rendering, which we named GeoTexDensifier. Our framework innovatively proposes a densification strategy that enforces the reconstructed Gaussians to more accurately comply with the actual geometric structure of the scene. Moreover, our densified Gaussian model contains a sufficient number of splats to support fully textured regions, while maintaining sparse Gaussian splat distribution in weakly textured areas. In these ways, our novel 3DGS framework is able to produce high-quality 3D Gaussian models, as shown in the comparative example of ``Church'' in Fig.~\ref{fig:teaser}. The original 3DGS~\cite{kerbl20233d} has ``over-reconstruction'' problem that causes missing details in textured regions such as the carpet, while state-of-the-art (SOTA) works like GeoGaussian~\cite{li2024geogaussian} improves the spatial structure of splat distribution and Pixel-GS~\cite{zhang2024pixel} further densifies the splats in some textured regions like the carpet, but both of them still lack sufficient splat sampling to support texture details. Mini-Splatting~\cite{fang2024mini} has sufficiently sampled splats but introduces over-densification situation in textureless regions, which might produce extra noisy Gaussians due to underconstrained visual ambiguity. In comparison, our GexTexDensifier delivers more accurate spatial distribution of Gaussian splat with less noise and more photorealistic novel view rendering effects according to the quantitative evaluation of the rendering results in the ground truth (GT) testing views on metrics of Structural SIMilarity (SSIM), Peak Signal to Noise Ratio (PSNR) in dB, and Learned Perceptual Image Patch Similarity (LPIPS).
Our GeoTexDensifier framework carries out a geometry-aware splitting strategy to guide the splitting position sampling to fit the actual geometric surfaces of the scene with the help of normal prior, and filter out noisy splats with improper initial positions according to our \textit{Validation of Depth Ratio Change} checking on relative depths provided by monocular depth prior. Meanwhile, a texture-aware densification strategy is adopted as auxiliary supplement for splitting of additional large Gaussians to produce a denser splat distribution in highly textured regions while remaining sparse in weakly textured areas, thereby preserving the overall high quality of Gaussian point cloud. The texture-aware densification and geometry-aware splitting strategies are fully integrated
in the iterative optimization to get the final high-quality photorealistic 3DGS model. 
Experiments on Mip-NeRF 360~\cite{barron2022mip}, Tanks and Temples~\cite{knapitsch2017tanks} datasets and self-captured scenes verify the
effectiveness and robustness of our GeoTexDensifier pipeline.

In summary, our GeoTexDensifier framework contributes in the following main aspects:
\begin{itemize}
\item We innovatively propose a geometry-aware splitting strategy that takes normal and relative depth priors to more reasonably guide the position sampling of the split Gaussians and eliminate the improperly sampled splats whose positions are far from the real surfaces, to ensure well distributed splats which comply with the actual geometric structure of the scene.
\item A texture-aware densification strategy is adopted as an auxiliary service for finding more contributive large splats in fully textured areas to be further split to fit texture details while keeping Gaussians sparse in weakly textured regions to maintain a high-quality spatial distribution of the recovered splats.
\item Our GeoTexDensifier framework iteratively optimizes the distribution and parameters of Gaussians under the combination action of texture-aware densification and geometry-aware splitting
to get photorealistic final 3DGS models with the best quality compared to SOTA works.
\end{itemize}

This paper is organized as follows. Section~\ref{sec:related-work} presents works related to our approach. Section~\ref{sec:system-overview} gives an overview of the proposed GeoTexDensifier framework. Section~\ref{sec:preliminary} briefly reviews the strategies of original 3DGS~\cite{kerbl20233d} and Mini-Splatting~\cite{fang2024mini}. The texture-aware densification, geometry-aware splitting modules are described in sections~\ref{sec:tex-aware-densification} and \ref{sec:geo-aware-splitting} respectively. Finally, we evaluate our GeoTexDensifier pipeline in section~\ref{sec:experiments}.

\section{Related Work}
\label{sec:related-work}
\subsection{Novel View Synthesis}
This paper studies Novel View Synthesis (NVS) field, which aims to generate realistic images of objects or scenes from unobserved viewpoints. NeRF~\cite{mildenhall2021nerf} has become a standard work in this field, modeling 3D scenes as continuous functions of density and color using a large Multi-Layer Perceptron (MLP) network, which enables view generation through volume rendering but at a high computational cost for both training and rendering. Various advancements~\cite{barron2021mip,barron2022mip,muller2022instant,sun2022direct,xu2022point,barron2023zip,chen2023neurbf,kulhanek2023tetra} aims at improving NVS quality and efficiency for training and perception. For example, Instant-NGP~\cite{muller2022instant} proposed a multi-resolution hash encoding for automatic detail focusing and reduced computational cost. Mip-NeRF~\cite{barron2021mip} replaced the point sampling mode with conical frustums and integrated positional encoding to address resolution-induced aliasing. The subsequent works Mip-NeRF 360~\cite{barron2022mip} and Zip-NeRF~\cite{barron2023zip} were designed to handle unbounded scenes and to be compatible with grid-based representations, respectively. However, the costly MLP perception and implicit volumetric representation remain significant challenges for NeRF model to be time and memory efficiently navigated or edited for practical applications.

Recently, 3DGS~\cite{kerbl20233d} introduced a new approach for NVS, by utilizing splatting-based rasterization to project anisotropic 3D Gaussians onto 2D screen, and calculating pixel colors through depth sorting and $\alpha$-blending. This splatting approach avoids complex ray marching to effectively enable real-time rendering for large-scale scenes, based on which several variants have emerged to further improve reconstruction scale~\cite{lin2024vastgaussian,liu2024citygaussian}
and enhance photorealism~\cite{yan2024multi,jiang2024gaussianshader,gao2023relightable,yu2024mip}. For example, both VastGaussian~\cite{lin2024vastgaussian} and CityGaussian~\cite{liu2024citygaussian} address reconstruction and rendering of large-scale scenes, with the latter further optimizing the training approach and Level-of-Detail (LOD) strategy for more efficient rendering. Relightable 3D Gaussian~\cite{gao2023relightable} additionally introduces normals, BRDF parameters, and direction-dependent incident lighting for photorealistic relighting. During the Gaussian optimization process, Adaptive Density Control (ADC) plays a crucial role by determining where to expand or shrink the spatial distribution of Gaussian points, particularly in ``under-reconstruction'' and ``over-reconstruction'' regions. However, due to possibly noisy initialization and insufficient geometric constraints, the growth of Gaussian splats can be undesirable particularly in textureless regions, leading to blur and artifacts in the rendered images. 
Some recent works have focused on this challenging problem, by proposing more reasonable densification strategies or introducing geometric priors for better ADC guidance to achieve better rendering results, which are discussed in detail in the following subsections.

\subsection{Gaussian Densification Strategy}
In the original 3DGS work~\cite{kerbl20233d}, the growth of Gaussians is determined by the magnitude of the average positional gradient. However, large Gaussian splats in over-reconstructed regions often provide limited gradient values to support their splitting. To increase the likelihood of splitting in these regions, some approaches have modified the densification criteria. Bul{\`o} et al.~\cite{bulo2024revising} designed a auxiliary per-pixel error function as densification criteria, rather than relying solely on positional gradients. Pixel-GS~\cite{zhang2024pixel} accounts for the maximal number of pixels each Gaussian contributes to different views as a compensational criteria to dynamically encourage splitting of large Gaussians. FreGS~\cite{zhang2024fregs} employs progressive frequency regularization to increase the average pixel gradient for coarse-to-fine Gaussian densification. Some other approaches attempt to reorganize the scene structure
to improve densification. For example, FSGS~\cite{zhu2023fsgs} organizes the scene into a graph structure based on proximity scores and defines new Gaussians at the edge centers, thereby enhancing control over ADC. Scaffold-GS~\cite{lu2024scaffold} searches anchors to establish a hierarchical and region-aware scene representation for constraining the spatial splat distribution to avoid free drifting and splitting of Gaussians. GeoGaussian~\cite{li2024geogaussian} detects
smoothly connected areas from input point cloud and initializes each point as a thin ellipsoid aligned with smooth surfaces to enhance densification.
Mini-Splatting~\cite{fang2024mini} introduces blur split and depth reinitialization as densification strategies for facilitating a uniform spatial distribution of splats.
These improved densification criterion commonly serve much like a compensational strategy to enrich the splats generated by the original work, while our GeoTexDensifier framework is an innovative strategy to make full use of geometry priors and texture gradients to guide our densification to a more accurate splat distribution.

\subsection{Gaussian Optimization with Geometry Prior}
Original 3DGS expects Gaussian splats expected to grow along the real scene surfaces, which is however not always conformed to. Consequently, some approaches have attempted to optimize Gaussians by incorporating geometric priors like depths and normals. Similar to MonoSDF~\cite{yu2022monosdf}, DN-Splatter~\cite{turkulainen2024dn} employs mono-depth and mono-normal estimates to constrain the rendering loss. DNGaussian~\cite{li2024dngaussian} introduces Hard and Soft Depth Regularization to optimize the positions and opacities of Gaussians, and employs Global-Local Depth Normalization to mitigate the sensitivity of scale-invariant loss to small depth errors. PGSR~\cite{chen2024pgsr} optimizes Gaussians by rendering unbiased depths combined with single-view and multi-view geometric consistency losses. Besides applying depth and normal constraints, both GaussianRoom~\cite{xiang2024gaussianroom} and GSDF~\cite{yu2024gsdf} utilize a learnable neural SDF field to guide the growth of splats, so as to simultaneously optimize both SDF and Gaussians. These geometry prior based methods rely on the geometric accuracy of the prior guidance, and are easily influenced by the incorrect depths or normals, especially when the truth depth scale is unavailable, while our approach only uses relative depth ratios for verifying newly generated splats, so as to relax the requirement on depth or scale accuracy.

\section{System Overview}
\label{sec:system-overview}
\begin{figure*}[tbh]%
\centering
\includegraphics[width=0.75\linewidth]{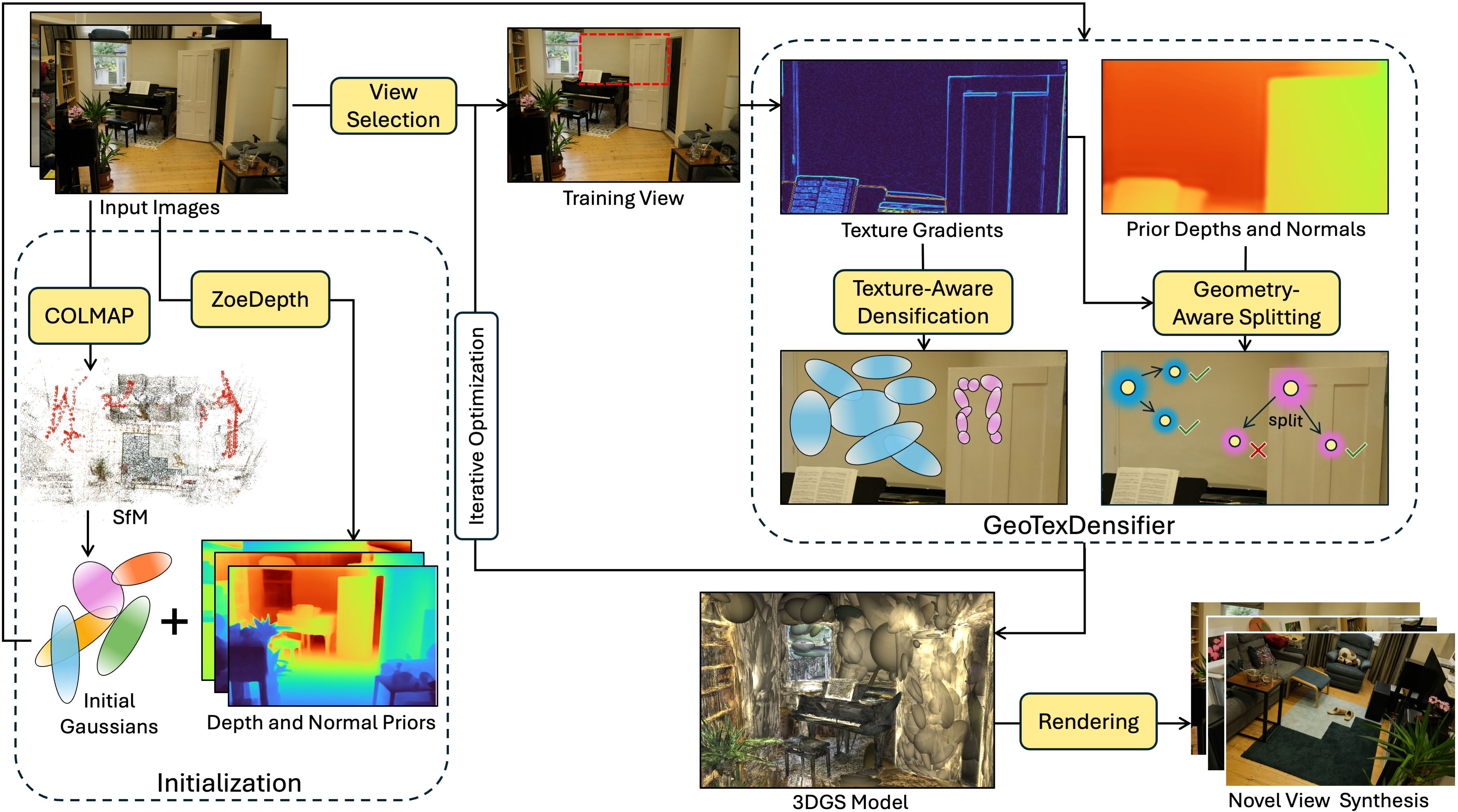}
\caption{System overview of our GeoTexDensifier, which consists of a geometry-aware splitting strategy for guiding the positioning of newly generated splats by depth and normal priors, and a texture-aware densification module which finds more contributive splats in fully textured areas to be further split to refine texture details 
in iterative optimization to get a well-structured 3DGS model with photorealistic rendering results.}
\label{fig:system-overview}
\end{figure*}

Suppose each scene has a set of multi-view RGB images captured with digital cameras as input, denoted by $\mathcal{I} = \{ I_t | t = 1,\dots,M \}$, where $M$ is the number of input images. Our GeoTexDensifier system is applied to the multi-view images as training views to robustly reconstruct an high-quality 3DGS model of the captured scene, which we defined as $\mathcal{G}$. Fig. \ref{fig:system-overview} outlines the proposed high-quality 3DGS framework. In the initialization stage, COLMAP~\cite{schoenberger2016sfm} is carried out first to estimate camera poses of all the input views denoted as $\{ \mathbf{M}_t | t = 1,\dots,M \}$, with sparse Structure-from-Motion (SfM) feature map points used to initialize a set of Gaussian splats. Meanwhile, a depth map is estimated for each input RGB image by ZoeDepth~\cite{bhat2023zoedepth} with its normal map computed from depths, denoted as $\mathcal{D} = \{ D_t | t = 1,\dots,M \}$ and $\mathcal{N} = \{ N_t | t = 1,\dots,M \}$ respectively. After initialization finishes, the adaptive optimization stage follows to control the splat distribution by densification and pruning, while iteratively refining the Gaussian parameters to enforce their rasterized images to be consistent with the texture details observed in the training views. We use a texture-aware densification strategy to provide sufficient Gaussian splats in fully textured areas while maintaining the sparsity of Gaussians in regions with weak textures to ensure more accurate spatial distribution of Gaussian points. Additionally, the estimated depths are used as priors to guide the splitting of Gaussian splats to better conform to the actural tangential directions of the geometric surfaces in weakly textured regions and avoid the generation of isolated or scattered Gaussians caused by the ambiguity due to lack of visual features. The texture-aware densification approach together with the geometry-guided splitting strategy constitute our geometry-texture-aware densification framework for reconstructing high-quality photorealistic 3DGS model, which will be described in detail in the following sections.

\section{Preliminary}
\label{sec:preliminary}
Original 3DGS~\cite{kerbl20233d} explicitly represents the scene with a collection of anisotropic 3D Gaussians that retains the differential properties of volumetric representation while enabling real-time rendering through a tile-based rasterization. Each Gaussian splat $G_k$ initially derived from a sparse SfM feature point, is defined by attributes including mean position $\mu_k$, spherical harmonic (SH) coefficients ${\mathrm{c}}_k$ to model its view-dependent color, anisotropic covariance $\Sigma_k$, and opacity $\alpha_k$, with $G_k \sim \mathcal{N}(\mu_k, \Sigma_k)$.
To ensure the covariance matrix $\Sigma_k$ to be semi-positive definite, it is decomposed into a diagonal scaling matrix ${\mathrm{S}}_k = \operatorname{diag}([s_{1}\ s_{2}\ s_{3}]) \in {\mathbb{R}}^{3\times 3}$ and a rotation quaternion ${\mathrm{R}}_k = [r_{1}\ r_{2}\ r_{3}] \in \mathrm{SO}(3)$, as $\Sigma_k = {\mathrm{R}}_k {\mathrm{S}}_k {\mathrm{S}}_k^{\top} {\mathrm{R}}_k^{\top}$.

3DGS renders a novel-view image $n$ by $\alpha$-blending of $K$ depth-sorted splats for each pixel $\bar{{\mathbf{x}}}$ as follows:
\begin{equation}
\begin{array}{*{20}{l}}
&{\mathrm{C}}(\bar{{\mathbf{x}}})=\sum_{k=1}^{K} \mathrm{c}_{k} w_k(\bar{{\mathbf{x}}}) \prod_{j=1}^{k-1}\big(1-w_j(\bar{{\mathbf{x}}})) \\
&w_k(\bar{{\mathbf{x}}}) = \alpha_k \bar{G}_k(\bar{{\mathbf{x}}}),
\end{array}
\label{eq:original-3dgs}
\end{equation}
where ${\mathrm{C}}(\bar{{\mathbf{x}}})$ represents the color rendered at image pixel $\bar{{\mathbf{x}}}$,
and $w_k(\bar{{\mathbf{x}}})$ is the rendering weight for $\alpha$-blending.
$\bar{G}_k \sim \mathcal{N}(\bar{\mu}_k, \bar{\Sigma}_k)$ is the projected 2D Gaussian distribution of $G_k$ through a local affine approximation of perspective transformation $\mathbf{W}$, defined as $\bar{\mu}_k = {\mathbf{W}} \mu_k$ and $\bar{\Sigma}_k = [{\mathbf{J}} {\mathbf{W}} \Sigma_k {\mathbf{W}}^{\top} {\mathbf{J}}^{\top}]_{1:2,1:2}$, with $\mathbf{J}$ the Jacobian form of $\mathbf{W}$ and $[\cdot]_{1:2,1:2}$ taking the first two rows and columns as sub-matrix. Finally, by leveraging the differentiable rasterizer and comparing the rendered images to the training views, all attributes of the 3D Gaussians can be learned and optimized by minimizing the ${\mathcal{L}}_1$ loss combined with a D-SSIM loss.

During the optimization of Gaussian parameters, ADC is applied to populate Gaussian splats in the empty areas, which focuses mainly on incomplete regions with missing Gaussians defined as ``under-reconstruction'', and areas covered by large-sized Gaussians as ``over-reconstruction''. Gaussians with large average view-space positional gradients are to be densified, by Gaussian cloning in under-reconstructed regions and splitting large-variance splats in over-reconstructed places, to get a sufficient number of Gaussian splats for more complete reconstruction. However, even with this ADC strategy, there still might be under-densification situation in regions with full textures. Recent works such as Mini-Splatting~\cite{fang2024mini}, Pixel-GS~\cite{zhang2024pixel} and Bul{\`o} et al.~\cite{bulo2024revising} have discussed this densification limitation and proposed corresponding densification improvement strategies to alleviate the problem. For example, Mini-Splatting incorporates blur split strategy combined with depth reinitialization to densify Gaussian splat distribution, followed by a simplification technique to suppress the total number of points for a more efficient Gaussian representation. Specifically, for each image $I_t$, a set of Gaussians with large blurry areas are identified by:
\begin{equation}
\begin{array}{*{20}{l}}
&{\mathcal{G}}_b^t = \{ G_i | S_i^t > {\mathcal{T}} \} \\
&S_i^t = \sum_{\bar{{\mathbf{x}}} = (1,1)}^{(W,H)} \delta (\mathbf{I}_i(\bar{{\mathbf{x}}}) = \mathbf{I}_{max}(\bar{{\mathbf{x}}})).
\end{array}
\label{eq:blur-split}
\end{equation}
Here $S_i^t$ represents the maximal contribution area of Gaussian $G_i$ to $I_t$, with $\mathbf{I}_i(\bar{{\mathbf{x}}})$ denoting the projection index of $G_i$ at pixel $\bar{{\mathbf{x}}} \in I_t$ and $\mathbf{I}_{max}(\bar{{\mathbf{x}}}) = \mathop{\arg\max}\limits_{k} \ w_k(\bar{{\mathbf{x}}})$ defining the rendered index with the maximal weight contribution at $\bar{{\mathbf{x}}}$. Threshold $\mathcal{T} = \theta W H$, with $\theta$ a coefficiency on the image resolution of $I_t$ denoted as $(W, H)$. The selected blurry splats are then split according to the original splitting strategy in \cite{kerbl20233d} during the adaptive optimization process. Although this strategy can enrich texture details in the under-densified textured areas, it only counts the Gaussian contribution sizes without consideration of texture richness for each area, so that the textureless regions will also be split in the same way according to the blur split, which is why we propose our texture-aware strategy to take texture information into consideration for a better densification.

\section{Texture-Aware Densification}
\label{sec:tex-aware-densification}
\begin{figure}[htb!]
    \centering
    \includegraphics[width=0.85\linewidth]{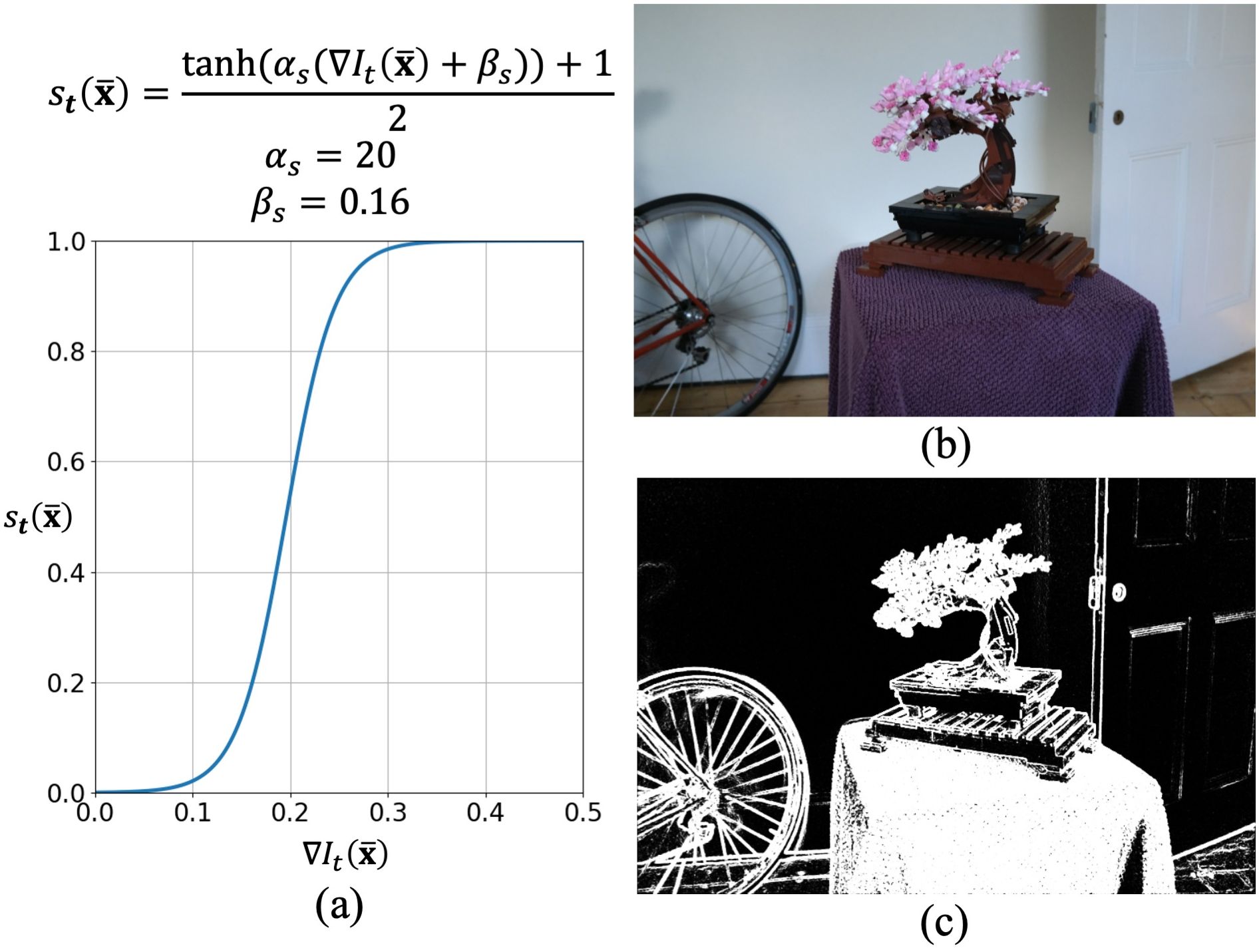}
    \includegraphics[width=0.85\linewidth]{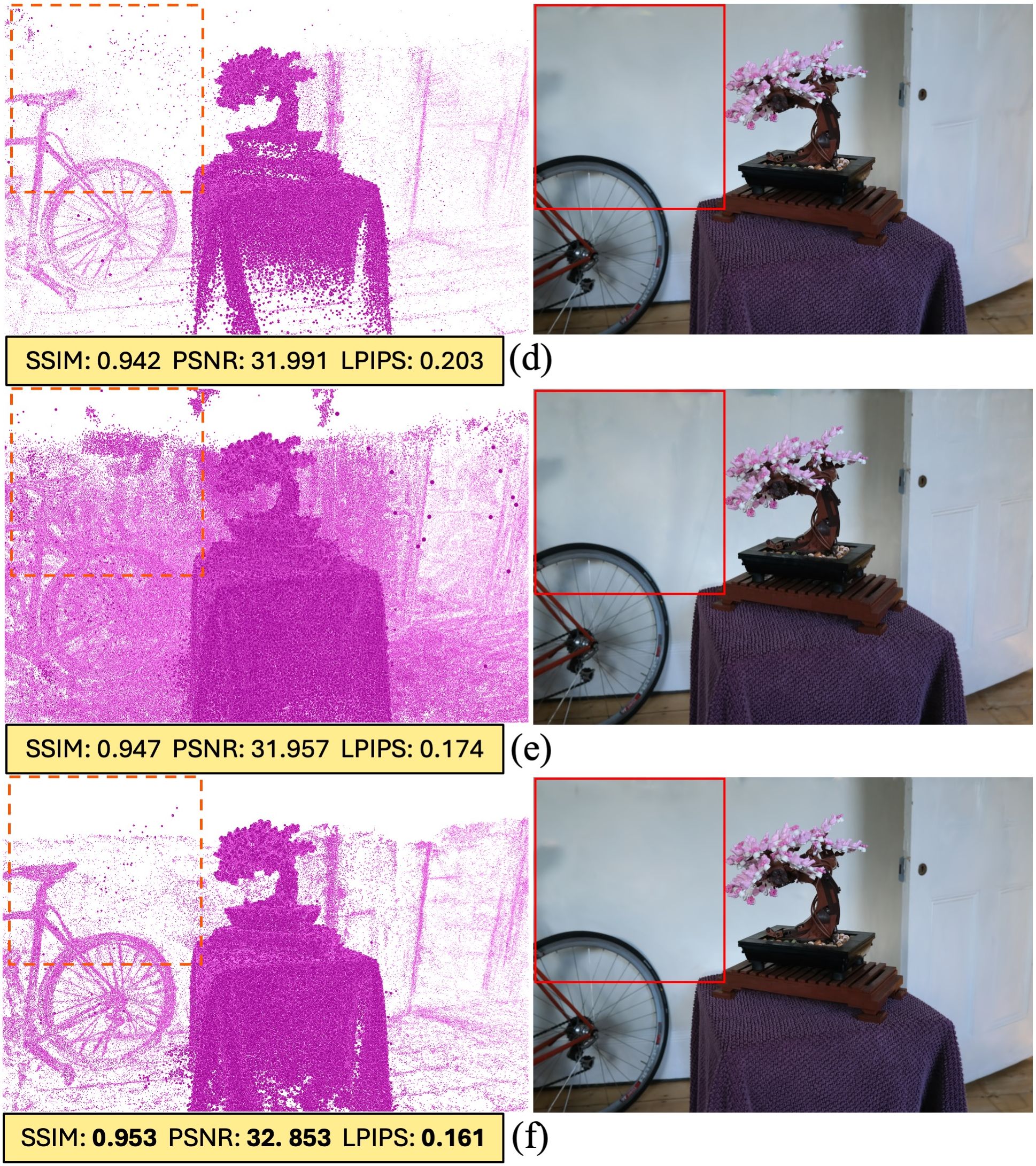}
    \caption{An example of texture-aware densification on case ``Bonsai'' from Mip-NeRF 360 dataset~\cite{barron2022mip}. (a) Activation function graph for $s_t(\bar{{\mathbf{x}}})$ by Eq.~\ref{eq:contribution-weight}. (b) A representative frame with its weight map computed by $s_t(\bar{{\mathbf{x}}})$ given in (c). (d)-(f) are the reconstructed Gaussian point clouds and rendered images by original 3DGS~\cite{kerbl20233d}, Mini-Splatting-D~\cite{fang2024mini} and our approach, respectively, to show both the well-structured Gaussian distribution and the best rendering quality of our densification strategy according to evaluation on SSIM, PSNR and LPIPS. }
\label{fig:texture-aware-weight}
\end{figure}
An ideal densification strategy is to maintain dense splat distribution in strongly textured areas, while still leaving Gaussian splats relatively sparse in more weakly textured regions such as surfaces with pure colors. The blur split strategy introduced by Mini-Splatting~\cite{fang2024mini} is adopted in our ADC process, which is although helpful for densification, but might cause the following side effects which do not actually conform to our ideal purpose: firstly, weakly textured regions often have Gaussians with large contribution areas, leading to over-splitting of splats to consequently generate too many inadequately optimized floaters which will influence the rendering results, and the second is that an excessive number of Gaussians in uniform splat distribution over the whole scene will cost too heavy memory consumption and large output file storage.

To address these issues, we propose a texture-aware densification approach, which fully leverages the texture richness of the training images to guide the densification degrees of Gaussian splats in regions of various textures, while controlling the growth of splats during the splitting process. In this way, our strategy ensures the densified splats more thoroughly optimized to improve the reconstruction quality. To better incorporate texture information as guidance, we designed a new weight $s_t(\bar{{\mathbf{x}}})$ for each pixel $\bar{{\mathbf{x}}} \in I_t$ to more reasonably count the contribution of each Gaussian $G_i$ for identifying blur splats in Eq.~\ref{eq:blur-split}. Activated by the texture gradient of each training image $I_t$, this new weight is calculated as follows:
\begin{equation}
    s_t(\bar{{\mathbf{x}}}) = \frac{\tanh (\alpha_s (\nabla I_t(\bar{{\mathbf{x}}}) + \beta_s)) + 1}{2},
\label{eq:contribution-weight}
\end{equation}
which effectively reflects the contribution to whether a Gaussian point should be split by a hyperbolic tangent function normalized to $[0, 1]$, which is activated by the image gradient whose gray values are normalized from $[0, 255]$ to $[0, 1]$, as shown in Fig.~\ref{fig:texture-aware-weight}(a). $\alpha_s$ and $\beta_s$ are the linear regression coefficiencies which we set to $20$ and $0.16$ respectively for our experiments.
Fig.~\ref{fig:texture-aware-weight}(c) shows a computed weight map by Eq.~\ref{eq:contribution-weight}. With the newly defined weight, we can redefine the maximal contribution area of each Gaussian $G_i$ to $I_t$ as $\hat{S}_i^t$, by modifying Eq.~\ref{eq:blur-split} to:
\begin{equation}
\begin{array}{*{20}{l}}
&{\mathcal{G}}_b^t = \{ G_i | \hat{S}_i^t > \hat{{\mathcal{T}}} \} \\
&\hat{S}_i^t = \sum_{\bar{{\mathbf{x}}} = (1,1)}^{(W,H)} s_t(\bar{{\mathbf{x}}}) \delta (\mathbf{I}_i(\bar{{\mathbf{x}}}) = \mathbf{I}_{max}(\bar{{\mathbf{x}}})),
\end{array}
\label{eq:texture-aware-densification}
\end{equation}
which collects statistics of pixel-wise texture gradients inside the contribution area of $G_i$ on $I_t$ instead of directly counting the contribution area $S_i^t$. In this way, fully textured contribution areas will have higher texture gradient statistics to consider the splats blurry and enforce them to be further split to get higher density of Gaussians, while weakly textured areas do not have sufficient texture gradient collection to support further splitting, so that their splat distributions are kept relatively sparse, as illustrated
in Fig.~\ref{fig:texture-aware-weight}(f).

The original threshold ${\mathcal{T}}$ in Eq.~\ref{eq:blur-split} is a fixed value.
For the redefined contribution area $\hat{S}_i^t$ calculated for each Gaussian splat, we further introduce an adaptive threshold $\hat{{\mathcal{T}}}$ to determine whether the Gaussian point is large enough to be split, which is defined as $\hat{{\mathcal{T}}} = {\mathcal{T}}_s + ({\mathcal{T}}_e - {\mathcal{T}}_s) \frac{l - l_s}{l_e - l_s}$,
where $l$ is the current iteration number, $l_s$ and $l_e$ are respectively the start and end iteration numbers for densification, and ${\mathcal{T}}_s$ and ${\mathcal{T}}_e$ are the start and end threshold values which we set to $40$ and $4$ respectively for our experiments.
Note that $\hat{{\mathcal{T}}}$ will decrease as the number of iterations increases,
to give the split Gaussians with smaller contribution areas
more chances to be further split and fully optimized in the next times of splitting.

Fig.~\ref{fig:texture-aware-weight}(d)-(f) gives an ablation study on case ``Bonsai'' to verify the usefulness of our texture-aware densification strategy. From the comparison of point clouds and rendered images by original 3DGS~\cite{kerbl20233d}, Mini-Splatting-D~\cite{fang2024mini} and our approach highlighted in the rectangles, we can see that our strategy more fully densifies the strongly-textured regions compared to original 3DGS, while keeping the weakly textured regions relatively sparse compared to the over-densified Mini-Splatting, to make a compact spatial distribution of splats that also performs the best in rendering quality.



\begin{table}[htb!]
\begin{minipage}{1.0\linewidth}
\centering
\caption{The statistics of time in minutes (m), GPU peak memory (G. P. M.) consumption (GB), and rendering frame rate of the reconstructed 3DGS model in Frames Per Second (FPS) on our pipeline and Mini-Splatting-D~\cite{fang2024mini} for the case ``Bonsai'' from Mip-NeRF 360~\cite{barron2022mip}.}
\label{tab:time-memory-fps}%
   \begin{tblr} {
     colspec = {l *{20}{c}}, 
     cell{1}{2} = {c=3}{},
     vline{2,5} = {-}{},
     hline{1,3,4} = {1-9}{},
     hline{2} = {2-9}{},
    }
    Methods              & Mini-Splatting-D & & & Ours \\
    Case                &  Time & G. P. M. & FPS & Time & G. P. M. & FPS \\
    Bonsai & 28m16s & 6.61GB & 132 & 42m31s & 3.75GB & 177
   \end{tblr}
\end{minipage}
\end{table}

This texture-aware densification serves as an auxiliary strategy that supplements the original splitting strategy for a better ADC. Table~\ref{tab:time-memory-fps} provides space/time resource statistics of training time, GPU peak memory (G. P. M.), and rendering frame rate (FPS) on our pipeline and Mini-Splatting-D~\cite{fang2024mini}, from which we can see that although our \textit{VDRC} checking and normal guidance strategies which are described in section~\ref{sec:geo-aware-splitting} increase our training time complexity, our texture-aware splitting successfully suppresses the GPU memory usage, resulting in a smaller number of Gaussians to be rendered in a higher FPS.
 
\section{Geometry-Aware Splitting}
\label{sec:geo-aware-splitting}
\begin{figure}[htb!]%
\centering
\includegraphics[width=0.99\linewidth]{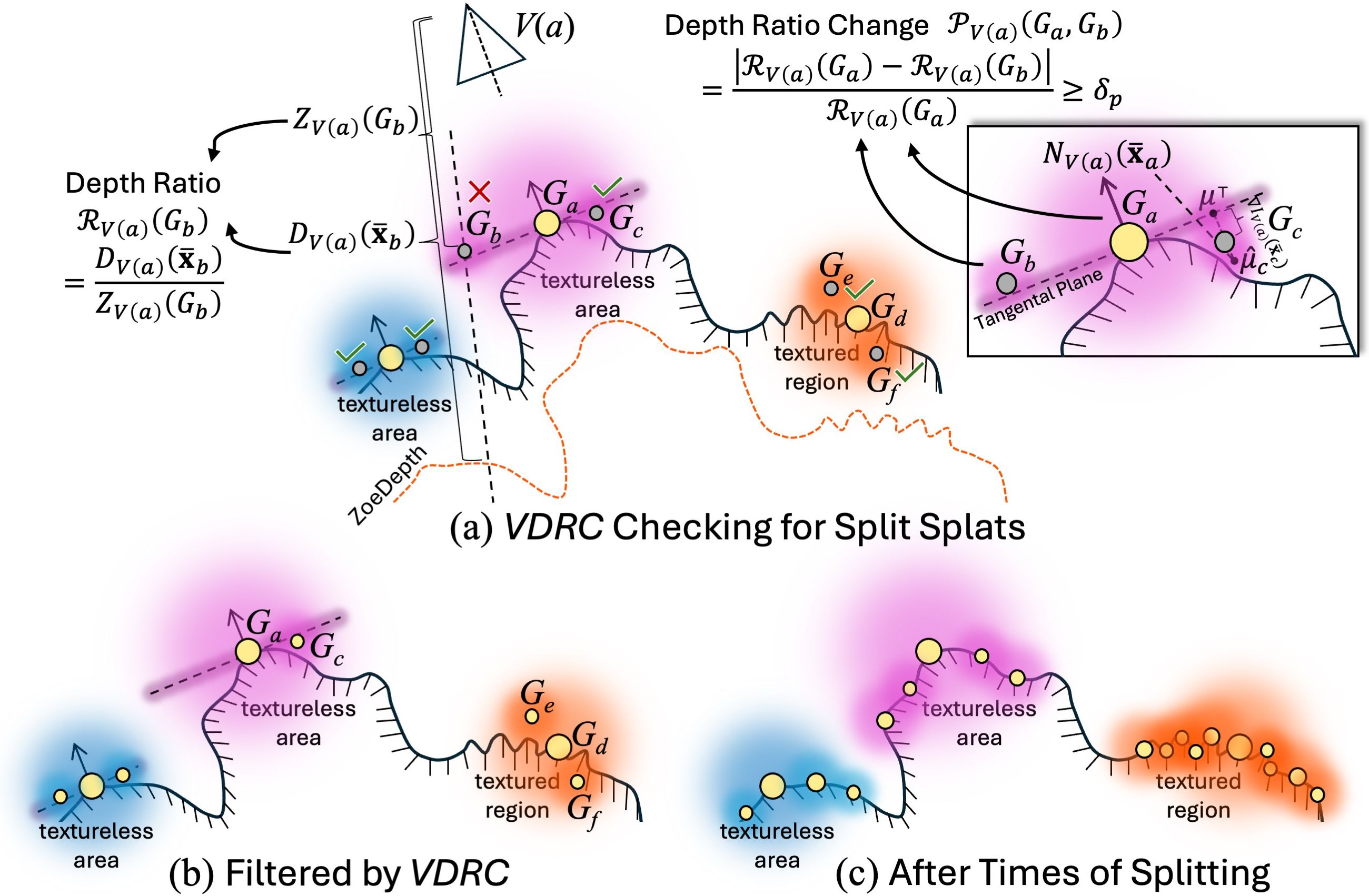}
\caption{A 2D simulation of our geometry-aware splitting. (a) Illustration of normal-guided splitting and \textit{VDRC} checking for children split from their parent splats. (b) The filtered splats by \textit{VDRC}, with noisy Gaussians eliminated and spatial distribution better fitting the actual structure. (c) The splats after several times of splitting to show the various distribution densities in textured and textureless regions.}
\label{fig:VDRC}
\end{figure}

\begin{figure*}[htb!]%
\centering
\includegraphics[width=0.83\linewidth]{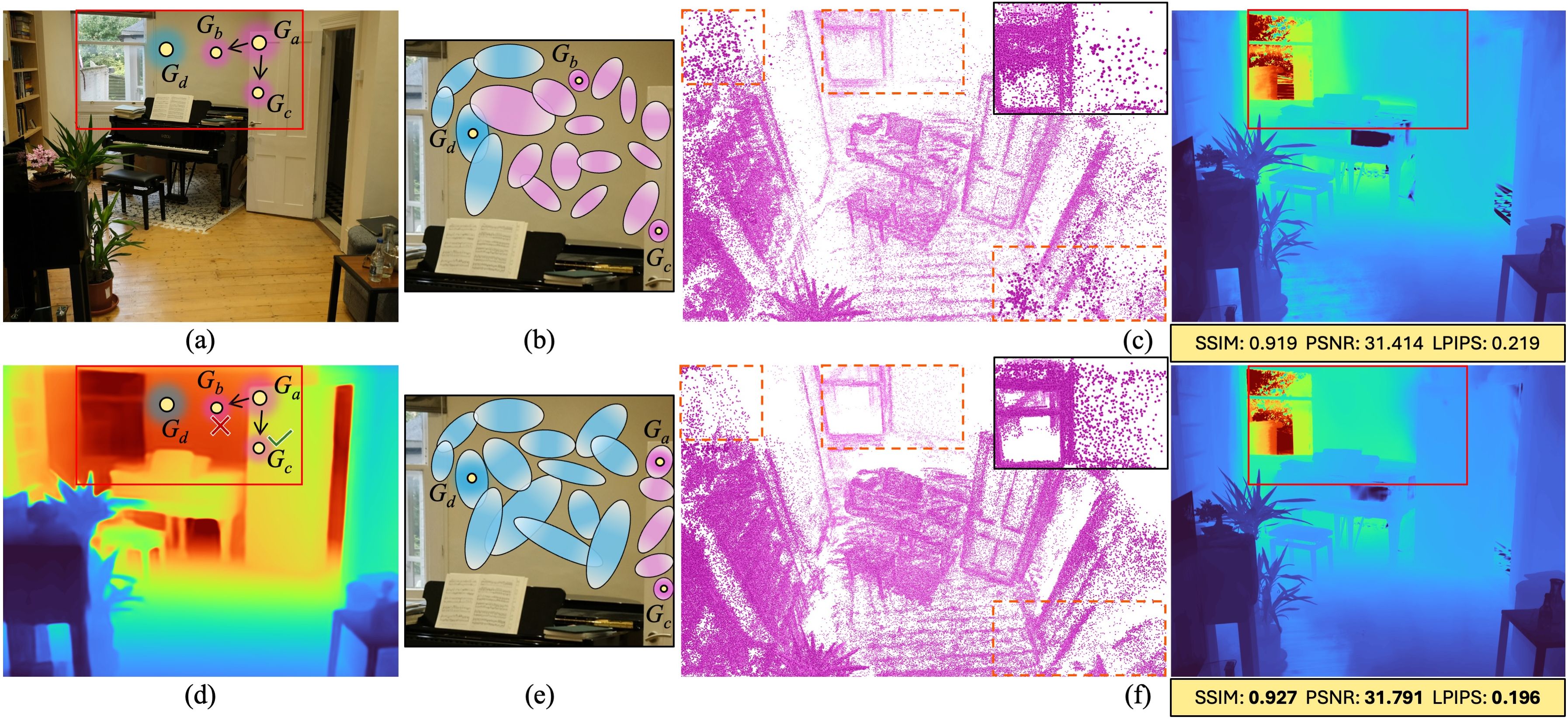}
\caption{\textit{VDRC} for case ``Room'' of Mip-NeRF 360 dataset~\cite{barron2022mip}. (a) A training view with two parent Gaussians denoted as $G_a$ and $G_d$, where $G_a$ attempts to be split into children $G_b$ and $G_c$. (b) Original 3DGS~\cite{kerbl20233d} advises $G_b$, $G_c$ and $G_d$ to be continuously split into new Gaussians, resulting in scattered splats due to the improper initial placement of $G_b$. (c) The positions of Gaussians and the rendered depth map by original 3DGS to show noisy splats of some highlighted places like the wall. (d) The predict depth map of (a) by ZoeDepth~\cite{bhat2023zoedepth}. (e) Our method discards $G_b$ by \textit{VDRC} with depth prior of (d) to make the new Gaussians split from $G_a$, $G_c$ and $G_d$.
(f) The splat positions and the rendered depth map by our method, with a more accurate geometric structure and fewer noisy points in places highlighted in red such as the wall.}
\label{fig:geo-split}
\end{figure*}

Besides using texture information to guide the distribution of Gaussian splats to fully textured areas, the initial placement of splats is also crucial for each time of splitting. Original 3DGS approach~\cite{kerbl20233d} splits an existing Gaussian into new ones located randomly within the elliptical splat scale with its Probability Density Function (PDF) as sampling guidance. These split splats can be further optimized to correct places near the real geometric surface if the regions contains plenty of textures observed by sufficient training views. However, there are usually large-sized regions with weak textures or without sufficient visible training views in real world scenarios, especially indoor scenes. If the initial 3D positions of the split splats seriously deviates from the real geometric surface,
it will be challenging for them to be refined to the right positions. This issue particularly occurs in textureless areas such as indoor walls, floors and doors, where the gradients of the loss function $\mathcal{L}_1$ is too small to move the split splats. Fig.~\ref{fig:geo-split}(a) gives an example of this issue: a parent Gaussian $G_a$ on a door generates a randomly split child Gaussian $G_b$ that attempts to fit the distant wall behind. However, $G_b$ is initially far from the wall's real place so that its position will not be optimized significantly to the right one. More seriously, $G_b$ will continue to be split into more incorrect splats which attempt to fit the wall in the next iterations, ultimately resulting in a scattered set of noisy splats,
as shown in Fig.~\ref{fig:geo-split}(b) and (c). Conversely, another split child Gaussian $G_c$ attempts to fit the door itself and is initially placed closer to it, which will not exhibit the same issue as $G_b$.

To better address this issue, we propose to leverage the depth maps of the training views estimated by ZoeDepth~\cite{bhat2023zoedepth} as an additional geometric prior for validating our splitting operations, considering the robustness of ZoeDepth to complicated senarios, especially scenes with weak textures. However, ZoeDepth predicts monocular depth maps without true scale of the captured scene. A straightforward idea is to align the depth map with the scale of sparse SfM points and check the discriminant validity of the initial placement of each newly split splat with the rescaled depth maps. Nevertheless, achieving pixel-level alignment of the monocular depth map is impractical due to the sparsity and noisy outliers of the SfM map points. Considering that the relative depths by SOTA monocular depth estimation networks like ZoeDepth are always reliable, we explore to directly utilize the relative depth information from the unscaled depth maps to achieve such discriminant checking. Consider the simulation example of Fig.~\ref{fig:VDRC}(a), where we aim to determine that the child $G_c$ is reasonable while the child $G_b$ is not. We compute the projection depth of the parent $G_a$ to its reference view $V(a)$ by $\mathrm{Z}_{V(a)}(G_a) = [\mathbf{M}_{V(a)} \mu_a]_z$ with $\mathbf{M}_{V(a)}$ its projection matrix, and the projection pixel $\bar{{\mathbf{x}}}_a = \pi(\mathbf{M}_{V(a)} \mu_a)$, with $\pi(x, y, z) = (f_u x / z + c_u, f_v y / z + c_v)$, in which $f_u$ and $f_v$ are the focal lengths in $uv$ directions of $V(a)$, and $(c_u, c_v)$ is the optical center, with its monocular depth obtained from ZoeDepth depth map as $D_{V(a)}(\bar{{\mathbf{x}}}_a)$. Then, a depth ratio is obtained by $\mathcal{R}_{V(a)}(G_a) = D_{V(a)}(\bar{{\mathbf{x}}}_a) / \mathrm{Z}_{V(a)}(G_a)$. Similarly, we calculate $\mathcal{R}_{V(a)}(G_b)$ and ${\mathcal{R}}_{V(a)}(G_c)$ respectively. After that, we compute a depth ratio change between $G_a$ and $G_c$ as:
\begin{equation}
{\mathcal{P}}_{V(a)}(G_a, G_c) = \frac{|{\mathcal{R}}_{V(a)}(G_a) - {\mathcal{R}}_{V(a)}(G_c)|} {{\mathcal{R}}_{V(a)}(G_a)}.
\label{eq:VDRC}
\end{equation}
It is evident that the depth ratio change between $G_a$ and $G_c$ is relatively small, whereas the relative depth ratio change ${\mathcal{P}}_{V(a)}(G_a, G_b)$ between $G_a$ and $G_b$ is large. We propose to discard the child splat whose relative depth ratio change from its parent splat exceeds a threshold $\delta_p$, to prevent generating an unsuitable child Gaussian like $G_b$. We set $\delta_p = 0.1$ for all our experiments. If $G_b$ is filtered out, $G_a$ will be reserved to take its place, as shown in Fig.~\ref{fig:VDRC}(b). We name this discriminant checking process as \textit{Validation of Depth Ratio Change} (\textit{VDRC}). As can be seen in Fig.~\ref{fig:geo-split}(e), thanks to the \textit{VDRC} checking to filter out noisy split splats, the splats on the front door will stay inside the door instead of being split back to the wall behind. With this \textit{VDRC} checking, our geometry-aware splitting is carried out whose details are given in the following subsections.

\subsection{Reference View Selection and Updating}
Intuitively, for each parent Gaussian $G_a$ to be split, we choose the view $V(a)$ where $G_a$ provides maximal rendering weight $w_{V(a)}(G_a)$ in $\alpha$-blending as its reference view. In our implementation, two additional attributes are added to each splat: an image id $V(a)$ and its rendering weight $w_{V(a)}(G_a)$
, which will be continuously updated during the training process.

Assuming that parent depth ratio ${\mathcal{R}}_{V(a)}(G_a)$ is reliable, we discard its child splat $G_b$ through \textit{VDRC} on condition that $\mathcal{P}_{V(a)}(G_a, G_b)$ is lower than a threshold which we set to $0.1$ for all our experiments, with the child depth ratio $\mathcal{R}_{V(a)}(G_b)$ calculated also on its parent's reference view $V(a)$ for fair comparison. As the position of $G_a$ continues to be optimized, its maximal rendering weight $w_{V(a)}(G_a)$ are recalculated and compared to the recorded one. If another view provides the maximal rendering weight, $V(a)$ and $w_{V(a)}(G_a)$ will be updated iteratively.

\begin{figure*}[htb!]
    \centering
    \includegraphics[width=0.95\linewidth]{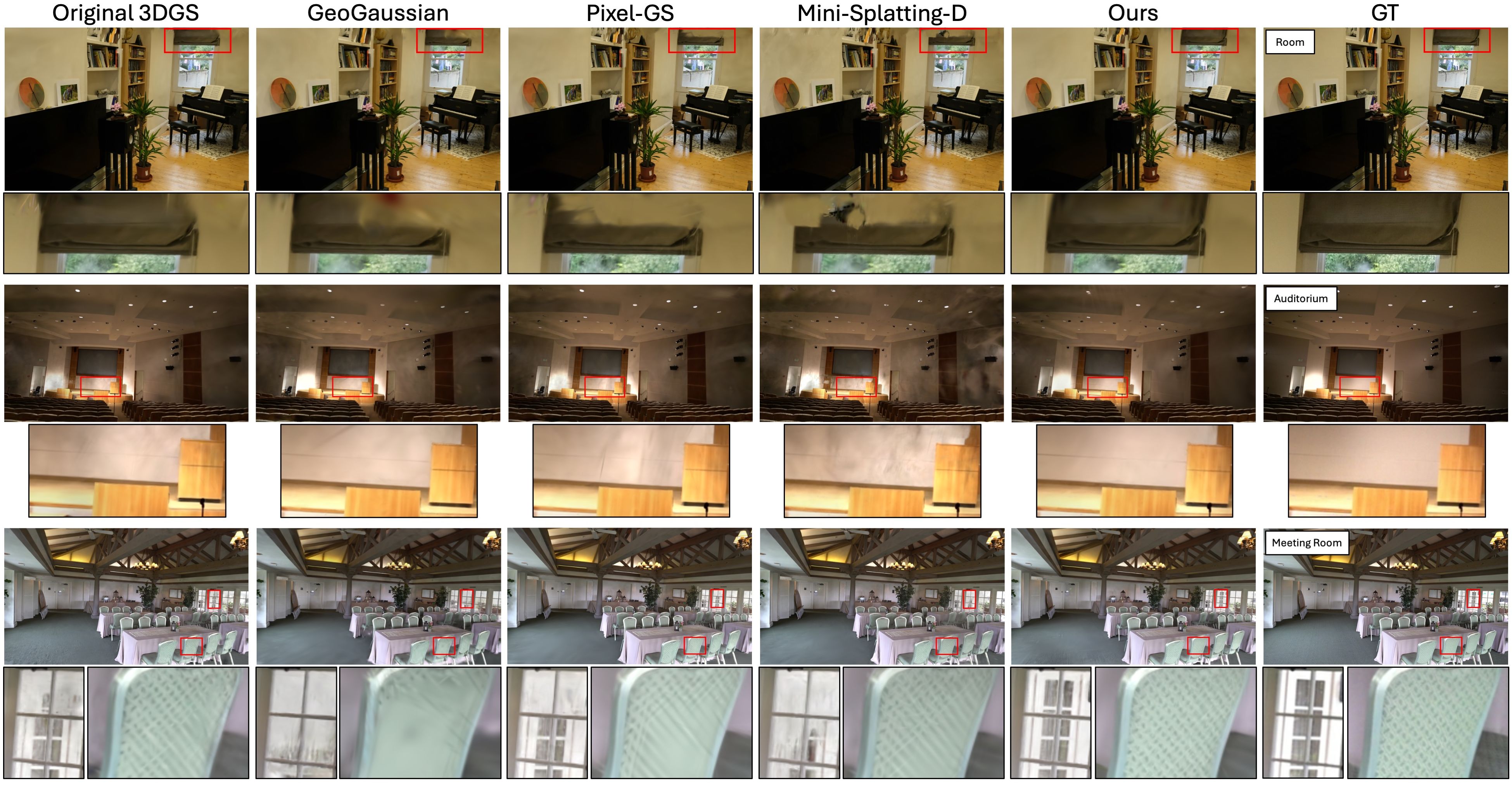}
    \includegraphics[width=0.95\linewidth]{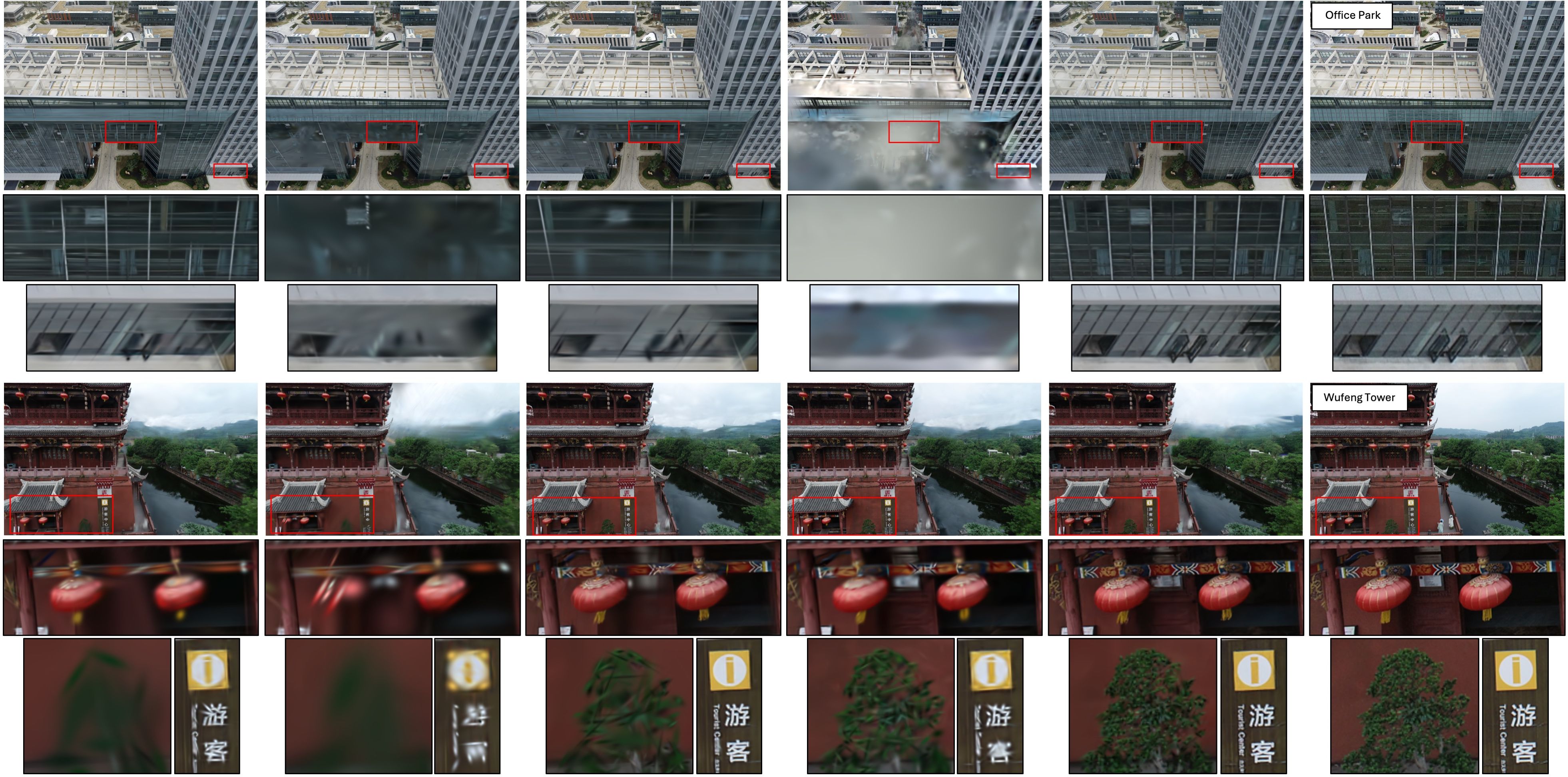}
    \caption{Qualitative comparison of our GeoTexDensifier pipeline to original 3DGS~\cite{kerbl20233d}, GeoGaussian~\cite{li2024geogaussian}, Pixel-GS~\cite{zhang2024pixel} and Mini-Splatting-D~\cite{fang2024mini} on the cases ``Room'', ``Auditorium'', ``Meeting Room'', ``Office Park'' and ``Wufeng Tower''. Some rendering details are highlighted in the rectangles to show the effectiveness of our proposed method in recovering the best texture details.}
\label{fig:qual-compare}
\end{figure*}

\begin{figure*}[htb!]
    \centering
    \includegraphics[width=0.8\linewidth]{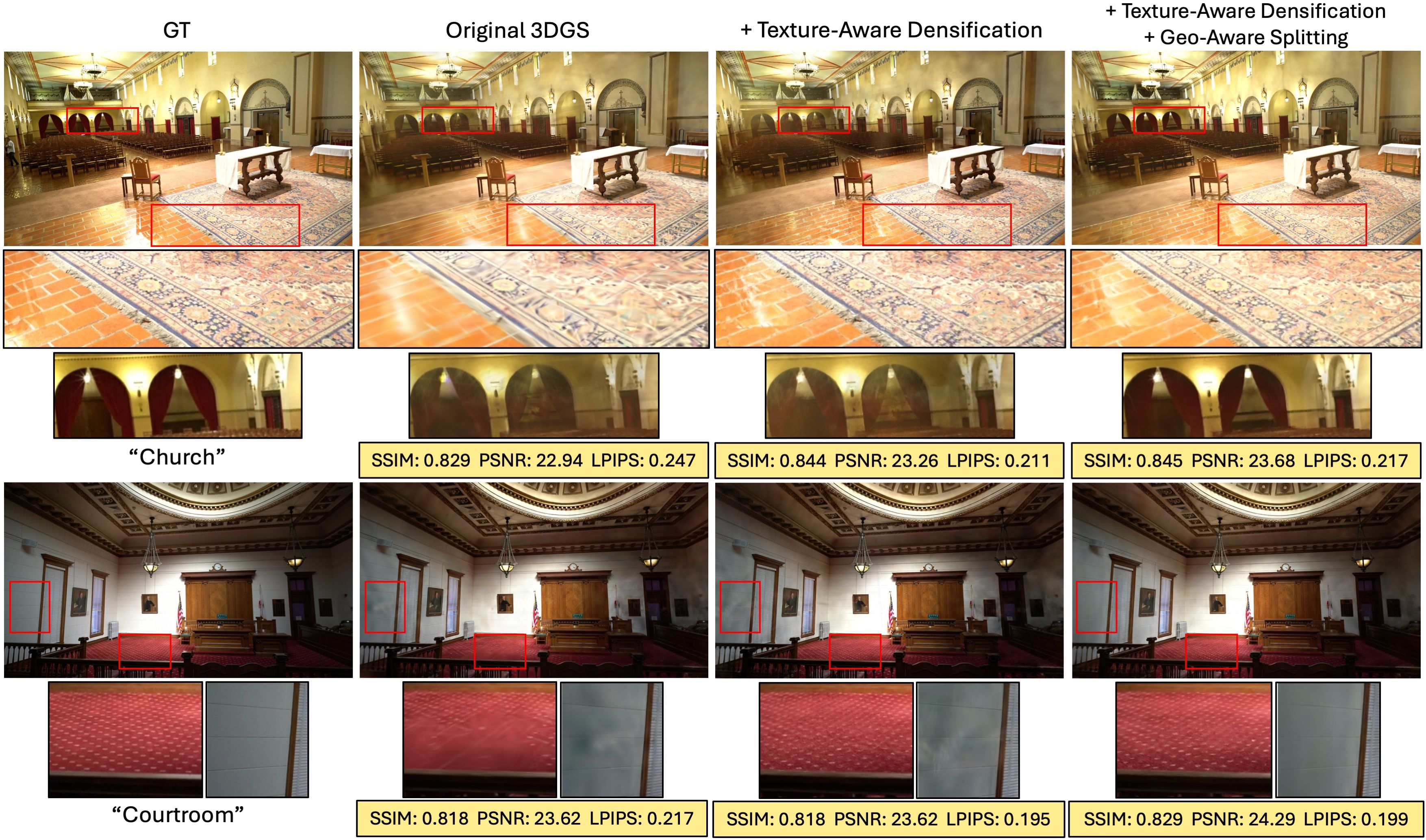}
    \caption{Exemplar effects of our texture-aware densification and geometry-aware splitting on two cases ``Church'' and ``Courtroom'' from Tanks and Temples dataset~\cite{knapitsch2017tanks}. We compare the results of original 3DGS~\cite{kerbl20233d}, original 3DGS combined with texture-aware densification and original one with both texture-aware densification and geometry-aware splitting, with metrics evaluated to show the effectiveness of the two proposed strategies in improving rendering quality.}
\label{fig:ablation-study}
\end{figure*}

\begin{figure}[htb!]
    \centering
    \includegraphics[width=0.95\linewidth]{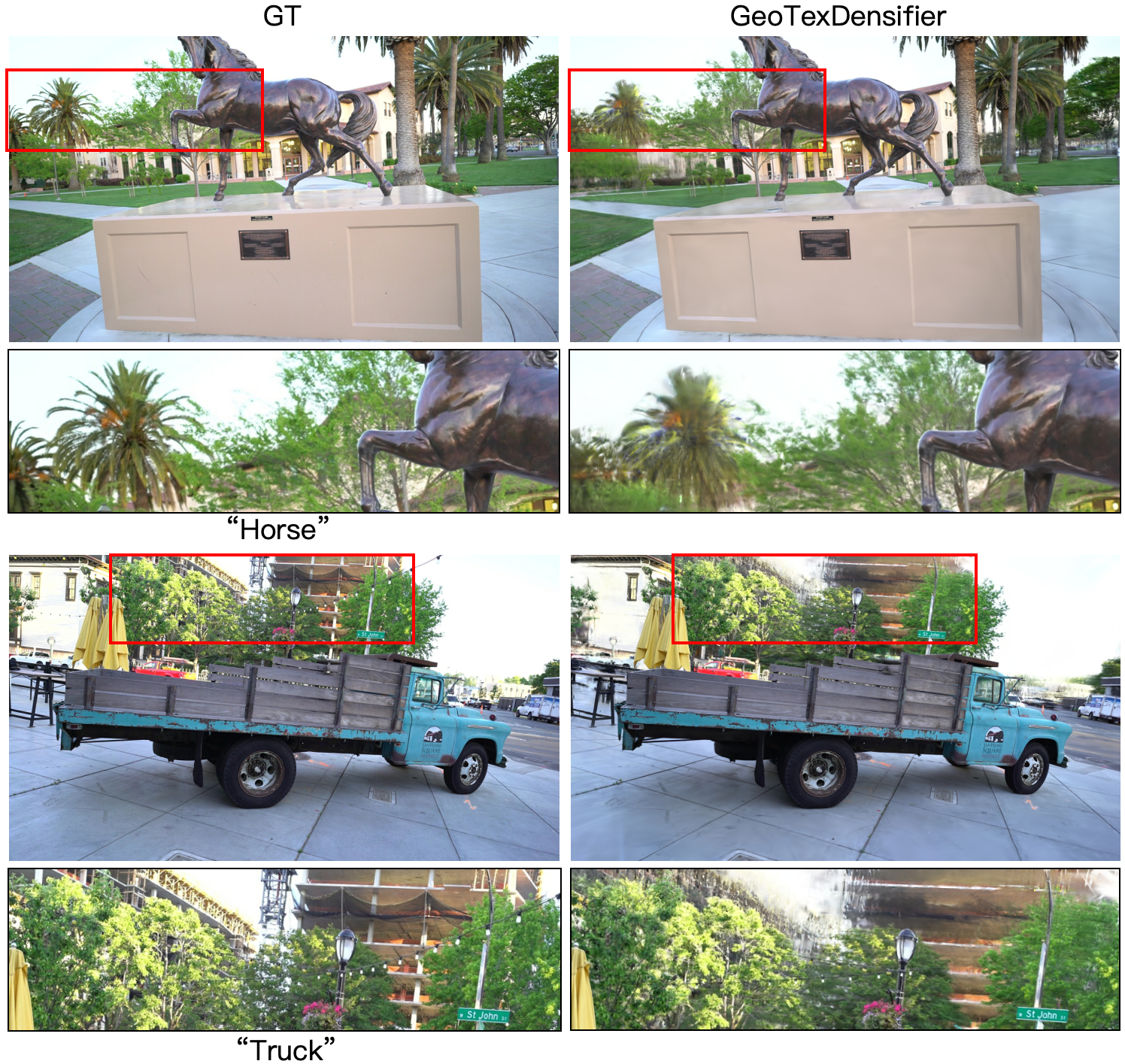}
    \caption{Two failure cases of ``Horse'' and ``Truck'' from Tanks and Temples dataset~\cite{knapitsch2017tanks}. Each case contains a representative GT testing view and the rendered 3D Gaussian splats by our GeoTexDensifier.}
\label{fig:failure}
\end{figure}

\begin{figure}[htb!]
    \centering
    \includegraphics[width=0.85\linewidth]{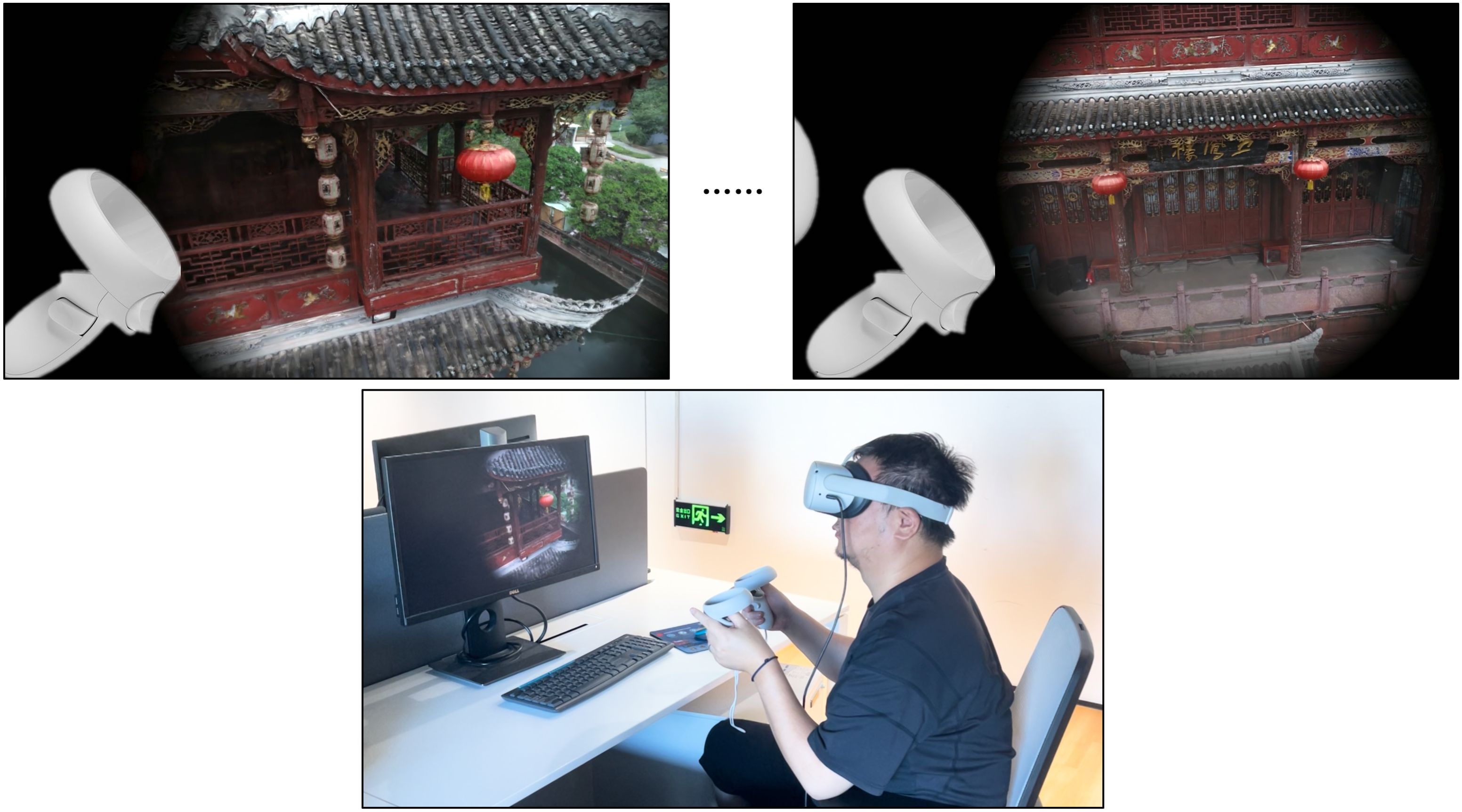}
    \caption{VR application of rendering and navigating the ``Wufeng Tower'' 3DGS model reconstructed by GeoTexDensifier on Gaussian Splatting VR Viewer Unity Plugin visualized on Oculus Quest II.}
\label{fig:VR-application}
\end{figure}

\subsection{\textit{VDRC} for Parent Splats}
Before each new time of splitting, a certain number of iterations might shift some parent splats to improper locations. For example, $G_a$ might move from the door towards the wall, becoming an unreliable Gaussian like $G_b$, and will continue to generate a series of improper splats which will all meet the condition of \textit{VDRC}. To better solve this issue, we apply \textit{VDRC} to the parent Gaussians before each time we split them. Since the initial Gaussians are inherited from the SfM map points, we filter out the map points whose track lengths are less than $3$ or reprojection errors exceed $1$ pixel, to ensure the correctness of the initial splat locations. We also choose the image view with the smallest reprojection error denoted by $\widehat{V}(a)$ as the initial reference view of each Gaussian $G_a$, and add two additional attributes to record $\widehat{V}(a)$ and its initial depth ratio $\widehat{{\mathcal{R}}}_{\widehat{V}(a)}(G_a)$. Before each time of splitting, each optimized parent Gaussian splat $G_a$ involved will be projected to its initial reference view $\widehat{V}(a)$ to calculate the current depth ratio ${\mathcal{R}}_{\widehat{V}(a)}(G_a)$, which is compared with its initial depth ratio through \textit{VDRC} to validate its own change ${\mathcal{P}}_{\widehat{V}(a)}(G_a) = |\widehat{{\mathcal{R}}}_{\widehat{V}(a)}(G_a) - {\mathcal{R}}_{\widehat{V}(a)}(G_a)| / \widehat{{\mathcal{R}}}_{\widehat{V}(a)}(G_a)$. Those splats not meeting the \textit{VDRC} requirement will be excluded from the following splitting process. If $G_a$ is reliable by self-validation, the splitting is carried out, and \textit{VDRC} continues to validate its two children $G_b$ and $G_c$ on their parent's reference view $V(a)$ by Eq.~\ref{eq:VDRC}. The valid child splat $G_c$ will regard its own initial reference view $\widehat{V}(c)$ inherited from $V(a)$ and initial depth ratio $\widehat{{\mathcal{R}}}_{\widehat{V}(c)}(G_c) = {\mathcal{R}}_{V(a)}(G_c)$, which will be further used to verify whether $G_c$ is to be divided into new Gaussians in the next time of splitting.

\subsection {Normal Guided Splitting}
Besides the use of monocular depths for checking validity of the split Gaussians, normal maps from ZoeDepth~\cite{bhat2023zoedepth} are also integrated as geometric guidance for splitting into more reasonable positions. For each validated child splat $G_c$, original random splitting within the elliptical scale of its parent $G_a$ will affect the smoothness and compact spatial distribution of Gaussian splats in weakly textured areas, as shown in the noisy Gaussian point cloud of Fig.~\ref{fig:geo-split}(c). We use normal maps combined with image textures to guide the initial placement of the newly split Gaussians. Also taking the simulation of Fig.~\ref{fig:VDRC}(a) as example, we first project $G_a$ and the randomly sampled $G_c$ to the reference view $V(a)$, where the projection pixels are denoted as $\bar{{\mathbf{x}}}_a$ and $\bar{{\mathbf{x}}}_c$ respectively. Then, with the normal $N_{V(a)}(\bar{{\mathbf{x}}}_a)$ at $\bar{{\mathbf{x}}}_a$ and the image gradient $\nabla I_{V(a)}(\bar{{\mathbf{x}}}_c)$ at $\bar{{\mathbf{x}}}_c$, an optimal position of $G_c$ can be determined as follows:
\begin{equation}
\begin{array}{*{20}{l}}
&\mu_c = (\widehat{\mu}_c - \mu^{\bot}) \nabla I_{V(a)}(\bar{{\mathbf{x}}}_c) + \mu^{\bot} \\
&\mu^{\bot} = \widehat{\mu}_c - ((\widehat{\mu}_c - \mu_a) \cdot N_{V(a)}(\bar{{\mathbf{x}}}_a)) N_{V(a)}(\bar{{\mathbf{x}}}_a),
\end{array}
\end{equation}
where $\widehat{\mu}_c$ represents the randomly sampled position of $G_c$, and $\mu^{\bot}$ is the perpendicular projection of $\widehat{\mu}_c$ onto the tangental plane of $G_a$ with $N_{V(a)}(\bar{{\mathbf{x}}}_a)$ as normal, whose plane equation can be represented as $(\mu_a, N_{V(a)}(\bar{{\mathbf{x}}}_a))$ in point-normal form, so that $N_{V(a)}(\bar{{\mathbf{x}}}_a) \cdot (\mu^{\bot} - \mu_a) = 0$. 
Since weakly textured regions lack sufficient visual details to enforce the split splats to move to the right positions, this normal-guided positioning enforces the child $G_c$ to stay close to the geometric surface of the parent $G_a$ in textureless areas with relatively small texture gradients. For fully textured areas, we believe that sufficient multi-view texture details are able to ensure the split children to be optimized to the correct places, so larger texture gradients tend to preserve the original randomness of the splat sampling instead of being excessively constrained by the geometric surface prior, as illustrated in the magnified region of Fig.~\ref{fig:VDRC}(a).

With normal guided splitting combined with \textit{VDRC} filtering, a better distributed Gaussian point cloud can be acquired that more accurately fits the actual scene structure with fewer noisy splats, as can be seen in the reconstructed ``Room'' point cloud of Fig.~\ref{fig:geo-split}(f) which also helps to produce better rendering quality evaluated by SSIM, PSNR and LPIPS.

\section{Experiments}
\label{sec:experiments}
\begin{table*}[htb!]
\begin{minipage}{1.0\linewidth}
\centering
\caption{Quantitative comparison of our method to original 3DGS~\cite{kerbl20233d}, original 3DGS with depth prior~\cite{kerbl20233d}, GeoGaussian~\cite{li2024geogaussian}, Pixel-GS~\cite{zhang2024pixel} and Mini-Splatting-D~\cite{fang2024mini} on Mip-NeRF 360 and Tanks and Temples datasets and two self-captured outdoor scenes, with the number of Gaussian splats denoted by $\lvert \mathcal{G} \rvert$ in millions, and evaluation of rendering quality metrics with SSIM, PSNR in dB and LPIPS.}
\label{tab:quant-compare}%
\resizebox{\linewidth}{!} {
\begin{tblr} {
  colspec = {l *{20}{c}}, 
  cell{1}{2} = {c=4}{},
  cell{1}{6} = {c=4}{},
  cell{1}{10} = {c=4}{},
  cell{1}{14} = {c=4}{},
  cell{1}{18} = {c=4}{},
  vline{2,6,10,14,18,22} = {-}{},
  hline{1,3,12,33,35} = {1-25}{},
  hline{2} = {2-25}{},
}
Methods              & Orig. 3DGS & & & & Orig. 3DGS + Depth Prior & & & & GeoGaussian & & & & Pixel-GS & & & & Mini-Splatting-D & & & & Ours \\
Cases                & $\lvert \mathcal{G} \rvert$ & SSIM & PSNR & LPIPS & $\lvert \mathcal{G} \rvert$ & SSIM & PSNR & LPIPS & $\lvert \mathcal{G} \rvert$ & SSIM & PSNR & LPIPS & $\lvert \mathcal{G} \rvert$ & SSIM & PSNR & LPIPS & $\lvert \mathcal{G} \rvert$ & SSIM & PSNR & LPIPS & $\lvert \mathcal{G} \rvert$ & SSIM & PSNR & LPIPS & \\
Bicyle (Mip-NeRF 360) & 6.1 & 0.765 & 25.205 & 0.21 & 5.24 & 0.763 & 25.216 & 0.215 & 2.8 & 0.75 & 24.891 & 0.237 & 9.06 & \SetCell{c3rd}{0.778} & \SetCell{c3rd}{25.265} & \SetCell{c3rd}{0.181} & 6.02 & \SetCell{c1st}{\bf{0.798}} & \SetCell{c1st}{\bf{25.542}} & \SetCell{c1st}{\bf{0.158}} & 6.03 & \SetCell{c2nd}{0.782} & \SetCell{c2nd}{25.388} & \SetCell{c2nd}{0.177} \\
Garden           & 5.88 & 0.866 & 27.34 & 0.107 & 4.12 & 0.864 & 27.39 & 0.109 &  2.91 &  0.858 & 27.117 & 0.121 & 8.78 & \SetCell{c2nd}{0.871} & \SetCell{c3rd}{27.492} & \SetCell{c2nd}{0.098} & 5.82 & \SetCell{c1st}{\bf{0.877}} & \SetCell{c2nd}{27.517} & \SetCell{c1st}{\bf{0.091}} & 5.36 & \SetCell{c3rd}{0.87} & \SetCell{c1st}{\bf{27.561}} & \SetCell{c3rd}{0.104} \\
Stump            & 4.92 & 0.773 & 26.621 & 0.215 & 4.42 & 0.77 & 26.593 & 0.217 & 1.93 & 0.753 & 25.502 & 0.247 & 6.66 & \SetCell{c2nd}{0.786} &\SetCell{c3rd}{26.842}  & \SetCell{c2nd}{0.187} & 5.37 & \SetCell{c1st}{\bf{0.804}} & \SetCell{c1st}{\bf{27.114}} & \SetCell{c1st}{\bf{0.169}} & 4.93 & \SetCell{c2nd}{0.786} & \SetCell{c2nd}{26.874} & \SetCell{c3rd}{0.195}\\
Treehill       & 3.77 & 0.633 & \SetCell{c2nd}{\bf{22.523}} & 0.325 &  3.37 & 0.634 & \SetCell{c1st}{\bf{22.525}} & 0.325 & 1.91 & 0.629 & 22.058 & 0.344 & 7.97 & \SetCell{c3rd}{0.634} & 22.21 & \SetCell{c2nd}{0.276} & 4.86 & \SetCell{c1st}{\bf{0.641}} & 22.222 & \SetCell{c1st}{\bf{0.261}} & 3.73 & \SetCell{c2nd}{0.635} & \SetCell{c3rd}{22.379} & \SetCell{c3rd}{0.293} \\
Flowers       & 3.63 & 0.606 & \SetCell{c2nd}{\bf{21.578}} & 0.336 & 3.22 & 0.606 & \SetCell{c1st}{\bf{21.582}} & 0.338 & 2.09 & 0.578 & 21.065 & 0.371 & 7.48 & \SetCell{c2nd}{0.635} & \SetCell{c3rd}{21.546} & \SetCell{c3rd}{0.262} & 4.87 & \SetCell{c1st}{\bf{0.643}} & 21.528 & \SetCell{c1st}{\bf{0.254}} & 6.18 & \SetCell{c3rd}{0.621} & 21.251 & \SetCell{c2nd}{0.26} \\
Room        & 1.54 & 0.919 & 31.414 & 0.219 & 1.37 & 0.919 & \SetCell{c2nd}{30.982} & 0.218 & 0.62 & 0.901 & 29.908 & 0.251 &2.57 & \SetCell{c3rd}{0.922} & \SetCell{c3rd}{31.579} & \SetCell{c3rd}{0.208} & 3.93 & \SetCell{c1st}{\bf{0.928}} & 31.576 & \SetCell{c1st}{\bf{0.187}} & 2.55 & \SetCell{c2nd}{0.927} & \SetCell{c1st}{\bf{31.93}} & \SetCell{c2nd}{0.195} \\
Counter          & 1.21 & 0.909 & \SetCell{c3rd}{28.993} & 0.2 & 1.13 & 0.907 & 28.874 & 0.2 & 0.56 & 0.889 & 27.899 & 0.235 & 2.57 & \SetCell{c2nd}{0.915} & \SetCell{c2nd}{29.186} & \SetCell{c3rd}{0.183} & 3.78 & \SetCell{c3rd}{0.913} & 28.728 & \SetCell{c1st}{\bf{0.171}} & 1.74  & \SetCell{c1st}{\bf{0.916}} & \SetCell{c1st}{\bf{29.216}} & \SetCell{c2nd}{0.179} \\
Bonsai           & 1.26 & 0.942 & 31.991 & 0.203 & 1.15 & 0.94 & \SetCell{c3rd}{32.219} & 0.204 & 0.76 & 0.925 & 31.026 & 0.225 & 2.08 & \SetCell{c2nd}{0.947} & \SetCell{c2nd}{32.554} & \SetCell{c3rd}{0.191} & 3.76 & \SetCell{c2nd}{0.947} & 31.957 & \SetCell{c2nd}{0.174} & 1.83 & \SetCell{c1st}{\bf{0.954}} & \SetCell{c1st}{\bf{32.952}} & \SetCell{c1st}{\bf{0.161}} \\
Kitchen         & 1.82 & 0.926 & 31.036 & 0.127 & 1.6 & 0.924 & 30.999 & 0.128 & 0.73 & 0.908 & 29.207 & 0.148 & 3.19 & \SetCell{c2nd}{0.931} & \SetCell{c2nd}{31.702} & \SetCell{c2nd}{0.119} & 3.67 & \SetCell{c1st}{\bf{0.933}} & \SetCell{c1st}{\bf{31.802}} & \SetCell{c1st}{\bf{0.114}} & 2.09 & \SetCell{c2nd}{0.931} & \SetCell{c3rd}{31.698} & \SetCell{c3rd}{0.12} \\
Courtroom (Tanks \& Temples)        & 2.99 & \SetCell{c2nd}{0.818} & 23.621 & \SetCell{c3rd}{0.217} & 3.23 & 0.806 & \SetCell{c2nd}{23.788} & 0.218 & 1.47 & 0.797 & 23.152 & 0.265 & 5.18 & \SetCell{c2nd}{0.818} & \SetCell{c3rd}{23.719} & \SetCell{c2nd}{0.215} & 5.24 & 0.792 & 21.73 & 0.219 & 4.78 & \SetCell{c1st}{\bf{0.829}} & \SetCell{c1st}{\bf{24.287}} & \SetCell{c1st}{\bf{0.199}} \\
Auditorium       & 0.69 & \SetCell{c3rd}{0.881} & 24.083 & \SetCell{c3rd}{0.247} & 0.77 & 0.871 & \SetCell{c2nd}{24.793} & \SetCell{c3rd}{0.264} & 0.25 & 0.867 & 23.765 & 0.292 & 1.21 & \SetCell{c2nd}{0.883} & \SetCell{c3rd}{24.351} & \SetCell{c2nd}{0.244} & 4.8 & 0.839 & 21.506 & 0.274 & 1.64 & \SetCell{c1st}{\bf{0.892}} & \SetCell{c1st}{\bf{25.24}} & \SetCell{c1st}{\bf{0.234}}  \\
Meetingroom
& 1.33 & \SetCell{c3rd}{0.88} & \SetCell{c3rd}{25.664} & 0.216
& 1.39 & 0.865 & \SetCell{c3rd}{25.699} & 0.227
& 0.57 & 0.861 & 24.752 & 0.264
& 3.16 & \SetCell{c2nd}{0.881} & \SetCell{c2nd}{25.857} & \SetCell{c2nd}{0.21}
& 4.21 & 0.869 & 24.859 & \SetCell{c2nd}{0.21}
& 2.2 & \SetCell{c1st}{\bf{0.888}} & \SetCell{c1st}{\bf{26.355}} & \SetCell{c1st}{\bf{0.201}} \\
Church           & 2.24 & \SetCell{c3rd}{0.829} & \SetCell{c3rd}{22.946} & 0.247 & 2.61 & 0.804 & 22.553 & 0.266 & 1 & 0.811 & 22.355 & 0.279 & 3.61 & \SetCell{c2nd}{0.833} & \SetCell{c2nd}{23.317} & \SetCell{c3rd}{0.243} & 4.86 & 0.818 & 21.89 & \SetCell{c2nd}{0.241} & 4.08 & \SetCell{c1st}{\bf{0.845}} & \SetCell{c1st}{\bf{23.678}} & \SetCell{c1st}{\bf{0.217}} \\
Ballroom     
& 3.42 &   0.807  &  \SetCell{c3rd}{23.902}  &  0.188
& 3.33 &  \SetCell{c2nd}{0.813} & \SetCell{c1st}{\bf{24.236}} & 0.183
& 1.23 & 0.788 & 23.31 & \SetCell{c1st}{\bf{0.141}}
& 6.07 & \SetCell{c3rd}{0.808} & 23.843  & 0.181
& 4.61 & 0.806 & 23.49  &  \SetCell{c3rd}{0.179}
& 4.45 & \SetCell{c1st}{\bf{0.815}} & \SetCell{c2nd}{23.99} & \SetCell{c2nd}{0.17} \\
Museum       
& 5.34 &  \SetCell{c3rd}{0.774}  &  \SetCell{c3rd}{21.22}   &   0.235 
& 4.49 &  \SetCell{c1st}{\bf{0.788}}  &  \SetCell{c1st}{\bf{21.956}}  &   \SetCell{c2nd}{0.222}
& 2.42 &  0.76   &  20.43   &   \SetCell{c1st}{\bf{0.176}}
& 9.62 &  0.769  &  21.15   &  0.239
& 6.09 &  0.77   &  20.624  &   0.231
& 4.76 &  \SetCell{c2nd}{0.778}  &  \SetCell{c2nd}{21.527}  &   \SetCell{c3rd}{0.226}  \\
Barn
& 1.03 & 0.817 & 26.828 & 0.259
& 1.13 & 0.815 & 26.921 & 0.256
& 0.39 & 0.788 & 25.54  & 0.24
& 4.22 & \SetCell{c3rd}{0.84} & \SetCell{c3rd}{27.234} & \SetCell{c3rd}{0.215}
& 3.88 & \SetCell{c1st}{\bf{0.862}} & \SetCell{c2nd}{27.455} & \SetCell{c1st}{\bf{0.175}}
& 4.02 & \SetCell{c2nd}{0.856} & \SetCell{c1st}{\bf{27.472}} & \SetCell{c2nd}{0.18} \\
Caterpillar
& 1.22  & 0.733 & \SetCell{c3rd}{22.738} & 0.3
& 1.52  & 0.74  & 22.608 & 0.292
& 0.42  & 0.692 & 21.45  & 0.297
& 3.96  & \SetCell{c2nd}{0.773} & \SetCell{c2nd}{23.051} & \SetCell{c3rd}{0.243}
& 3.9   & \SetCell{c2nd}{0.773} & 22.667 & \SetCell{c2nd}{0.232}
& 3.07  & \SetCell{c1st}{\bf{0.775}} & \SetCell{c1st}{\bf{23.185}} & \SetCell{c1st}{\bf{0.228}} \\
Courthouse
& 0.62 & 0.781 & 21.919 & 0.288
& 0.89 & \SetCell{c2nd}{0.79}  & \SetCell{c1st}{\bf{22.305}} & \SetCell{c2nd}{0.271}
& 0.2  & 0.755 & 20.57  & 0.306
& 1.36 & \SetCell{c3rd}{0.787} & \SetCell{c3rd}{22.126} & \SetCell{c3rd}{0.279} 
& 5.75 & 0.736 & 19.073 & 0.326
& 1.23 & \SetCell{c1st}{\bf{0.796}} & \SetCell{c2nd}{22.194} & \SetCell{c1st}{\bf{0.254}} \\
Family
& 2.31  & 0.844 & 24.024 & 0.191
& 2.23  & \SetCell{c2nd}{0.848} & \SetCell{c1st}{\bf{24.289}} & 0.182
& 1.12  & 0.807 & 22.460 & \SetCell{c2nd}{0.164}
& 3.86  & 0.847 & 23.942 & \SetCell{c3rd}{0.174}
& 4.21  & \SetCell{c1st}{\bf{0.856}} & \SetCell{c2nd}{24.27}  & \SetCell{c1st}{\bf{0.154}}
& 2.07  & \SetCell{c2nd}{0.848} & \SetCell{c3rd}{24.218} & 0.187 \\
Francis
& 0.81  & 0.881 & 27.188 & 0.265
& 0.87  & 0.883 & \SetCell{c1st}{\bf{27.798}} & 0.267
& 0.36  & 0.86  & 25.13 & \SetCell{c1st}{\bf{0.162}}
& 2.02  & \SetCell{c1st}{\bf{0.889}} & \SetCell{c2nd}{27.521} & 0.249
& 3.89  & \SetCell{c2nd}{0.888} & 26.203 & \SetCell{c2nd}{0.238}
& 1.27  & \SetCell{c2nd}{0.888} & \SetCell{c3rd}{27.375} & \SetCell{c3rd}{0.245} \\
Horse
& 1.32  & 0.866 & 23.481 & 0.187
& 1.49  & \SetCell{c1st}{\bf{0.874}} & \SetCell{c1st}{\bf{24.249}} & 0.175
& 0.72  & 0.85  & 22.73  & \SetCell{c1st}{\bf{0.139}}
& 2.39  & \SetCell{c3rd}{0.869} & \SetCell{c3rd}{23.744} & \SetCell{c3rd}{0.167}
& 3.91  & \SetCell{c2nd}{0.87}  & \SetCell{c2nd}{23.849} & \SetCell{c2nd}{0.152}
& 1.18  & 0.862 & 23.613 & 0.187 \\
Ignatius
& 3.45  & 0.732 & 21.245 & 0.264
& 2.96  & 0.736 & \SetCell{c1st}{\bf{21.817}} & 0.259
& 1.56  & 0.704 & 20.460 & \SetCell{c3rd}{0.234}
& 5.92  & \SetCell{c2nd}{0.744} & \SetCell{c3rd}{21.59}  & 0.235
& 4.75  & \SetCell{c1st}{\bf{0.747}} & 21.273 & \SetCell{c1st}{\bf{0.217}}
& 4.46 & \SetCell{c2nd}{0.744} & \SetCell{c2nd}{21.781} & \SetCell{c2nd}{0.228} \\
Lighthouse
& 0.85  & \SetCell{c2nd}{0.826} & \SetCell{c3rd}{21.966} & 0.234
& 0.97  & \SetCell{c1st}{\bf{0.83}}  & \SetCell{c1st}{\bf{22.612}} & \SetCell{c3rd}{0.224}
& 0.26  & 0.803 & 21.08  & \SetCell{c1st}{\bf{0.192}}
& 2.81  & 0.825 & 21.495 & 0.225
& 4.76  & 0.79  & 18.679 & 0.267
& 0.88  & \SetCell{c1st}{\bf{0.83}}  & \SetCell{c2nd}{22.434} & \SetCell{c2nd}{0.218} \\
M60
& 1.75  & 0.887 & 26.839 & 0.195
& 1.58  & 0.889 & \SetCell{c1st}{\bf{27.361}} & 0.193
& 0.75  & 0.87  & 26.040 & \SetCell{c1st}{\bf{0.151}}
& 5.04  & \SetCell{c1st}{\bf{0.896}} & \SetCell{c2nd}{27.329} & 0.164
& 4.14  & \SetCell{c2nd}{0.895} & 26.961 & \SetCell{c2nd}{0.153}
& 2.42  & \SetCell{c2nd}{0.895} & \SetCell{c3rd}{27.17} & \SetCell{c3rd}{0.163}  \\
Palace
& 0.66  & 0.741 & 19.936 & 0.359
& 0.84  & \SetCell{c3rd}{0.745} & \SetCell{c1st}{\bf{20.398}} & \SetCell{c2nd}{0.342}
& 0.23  & 0.725 & 19.51  & 0.358
& 1.90  & \SetCell{c2nd}{0.746} & \SetCell{c3rd}{20.014} & \SetCell{c3rd}{0.346}
& 7.28  & 0.666 & 15.467 & 0.444
& 0.85  & \SetCell{c1st}{\bf{0.751}} & \SetCell{c2nd}{20.352} & \SetCell{c1st}{\bf{0.326}} \\
Panther
& 1.96  & \SetCell{c3rd}{0.894} & \SetCell{c2nd}{27.734} & 0.185
& 1.73  & \SetCell{c3rd}{0.894} & \SetCell{c1st}{\bf{27.753}} & 0.185
& 0.96  & 0.882 & 27.04  & \SetCell{c1st}{\bf{0.122}}
& 5.14  & \SetCell{c1st}{\bf{0.899}} & \SetCell{c3rd}{27.705} & \SetCell{c2nd}{0.158}
& 4.41  & 0.871 & 25.571 & 0.184
& 1.79  & \SetCell{c2nd}{0.896} & 27.675 & \SetCell{c3rd}{0.177} \\
Playground
& 2.32  & 0.801 & 24.209 & 0.267
& 2.03  & 0.805 & 24.684 & 0.262
& 1.15  & 0.775 & 23.3   & 0.224
& 4.17  & \SetCell{c3rd}{0.832} & \SetCell{c3rd}{24.829} & \SetCell{c3rd}{0.217}
& 4.56  & \SetCell{c1st}{\bf{0.871}} & \SetCell{c1st}{\bf{25.571}} & \SetCell{c1st}{\bf{0.184}}
& 3.21  & \SetCell{c2nd}{0.835} & \SetCell{c2nd}{24.853} & \SetCell{c2nd}{0.197} \\
Temple
& 0.76  & 0.788 & \SetCell{c3rd}{20.021} & 0.292
& 1.03  & \SetCell{c1st}{\bf{0.804}} & \SetCell{c1st}{\bf{21.072}} & \SetCell{c2nd}{0.27}
& 0.29  & 0.772 & 19.630 & \SetCell{c2nd}{0.27}
& 3.38  & \SetCell{c3rd}{0.789} & 19.993 & 0.272
& 5.46  & 0.7   & 15.755 & 0.358
& 0.91  & \SetCell{c2nd}{0.795} & \SetCell{c2nd}{20.483} & \SetCell{c1st}{\bf{0.266}} \\
Train
& 0.93  & 0.762 & \SetCell{c2nd}{21.269} & 0.282
& 1.28  & \SetCell{c3rd}{0.771} & \SetCell{c1st}{\bf{21.438}} & 0.265 
& 0.32  & 0.728 & 20.05  & 0.27
& 3.5   & \SetCell{c2nd}{0.778} & 21.076 & \SetCell{c2nd}{0.25}
& 4.05  & 0.766 & 19.621 & \SetCell{c2nd}{0.25}
& 2.59  & \SetCell{c1st}{\bf{0.785}} & \SetCell{c3rd}{21.264} & \SetCell{c1st}{\bf{0.224}} \\
Truck
& 2.88  & 0.816 & 23.157 & 0.236
& 2.64  & \SetCell{c3rd}{0.819} & \SetCell{c1st}{\bf{23.5}}   & 0.227
& 1.2   & 0.791 & 22.18  & \SetCell{c1st}{\bf{0.155}}
& 6.23  & \SetCell{c2nd}{0.823} & \SetCell{c2nd}{23.314} & \SetCell{c3rd}{0.204}
& 4.64  & \SetCell{c1st}{\bf{0.83}}  & \SetCell{c3rd}{23.263} & \SetCell{c2nd}{0.165}
& 2.74  & \SetCell{c3rd}{0.819} & 23.199 & 0.226 \\
Wufeng Tower
& 4.69 & 0.775 & \SetCell{c3rd}{23.073} & 0.292
& 5.21 & \SetCell{c2nd}{0.781} & \SetCell{c2nd}{23.159} & 0.286
& 1.62 & 0.729 & 21.351 & 0.355
& 11.98 & \SetCell{c3rd}{0.777} & 22.848 & \SetCell{c2nd}{0.267}
& 5.86 & 0.768 & 22.495 & \SetCell{c3rd}{0.273}
& 10.34 & \SetCell{c1st}{\bf{0.8}} & \SetCell{c1st}{\bf{23.402}} & \SetCell{c1st}{\bf{0.232}} \\
Office Park
& 2.09 & \SetCell{c2nd}{0.684} & \SetCell{c3rd}{22.96} & \SetCell{c2nd}{0.42}
& 2.37 & 0.819 & 23.394 & 0.396
& 0.53 & 0.643 & 22.588 & 0.477
& 1.82 & \SetCell{c3rd}{0.668} & \SetCell{c2nd}{22.991} & \SetCell{c3rd}{0.448}
& 7.16 & 0.527 & 15.413 & 0.557
& 7.02 & \SetCell{c1st}{\bf{0.754}} & \SetCell{c1st}{\bf{24.309}} & \SetCell{c1st}{\bf{0.327}} \\
\end{tblr}
}
\end{minipage}
\end{table*}

\begin{table*}[htb!]
\begin{minipage}{1.0\linewidth}
\centering
\caption{Ablation comparisons of only geometry-aware splitting (Geo.), only texture-aware densification (Tex.), texture-aware densification with \textit{VDRC} (depth prior) (Tex. + VDRC), texture-aware densification with normal guidance (Tex. + Normal), and GeoTexDensifier (Tex. + Geo.) on Mip-NeRF 360~\cite{barron2022mip}, with the number of Gaussian splats denoted by $\lvert \mathcal{G} \rvert$ in millions, and evaluation of rendering quality metrics with SSIM, PSNR in dB and LPIPS.}
\label{tab:ablation}%
\resizebox{\linewidth}{!} {
\begin{tblr} {
  colspec = {l *{20}{c}}, 
  cell{1}{2} = {c=4}{},
  cell{1}{6} = {c=4}{},
  cell{1}{10} = {c=4}{},
  cell{1}{14} = {c=4}{},
  cell{1}{18} = {c=4}{},
  vline{2,6,10,14,18} = {-}{},
  hline{1,3,12} = {1-21}{},
  hline{2} = {2-21}{},
}
Methods              & Only Geo. (\textit{VDRC} + Normal) & & & & Only Tex. & & & & Tex. + \textit{VDRC} & & & & Tex. + Normal & & & & GeoTexDensifier (Tex. + Geo.) \\
Cases                & $\lvert \mathcal{G} \rvert$ & SSIM & PSNR & LPIPS & $\lvert \mathcal{G} \rvert$ & SSIM & PSNR & LPIPS & $\lvert \mathcal{G} \rvert$ & SSIM & PSNR & LPIPS & $\lvert \mathcal{G} \rvert$ & SSIM & PSNR & LPIPS & $\lvert \mathcal{G} \rvert$ & SSIM & PSNR & LPIPS & \\
Bicycle    & 5.49 & 0.771 & 25.262 & 0.199 & 6.28 & 0.777 & \SetCell{c3rd}{25.372} & 0.19 & 5.73 & \SetCell{c2nd}{0.78} & 25.367 & \SetCell{c2nd}{0.186} & 5.81 & \SetCell{c2nd}{0.78} & \SetCell{c1st}{\bf{25.407}} & \SetCell{c3rd}{0.187} & 6.03 & \SetCell{c1st}{\bf{0.782}} & \SetCell{c2nd}{25.388} & \SetCell{c1st}{\bf{0.177}} \\
Garden    & 5.33 & 0.868 & 27.433 & 0.105 & 5.91 & 0.869 & 27.536 & 0.105 & 5.36 & \SetCell{c1st}{\bf{0.87}} & \SetCell{c3rd}{27.541} & \SetCell{c1st}{\bf{0.103}} & 5.37 & \SetCell{c1st}{\bf{0.87}} & \SetCell{c2nd}{27.555} & \SetCell{c1st}{\bf{0.103}} & 5.36 & \SetCell{c1st}{\bf{0.87}} & \SetCell{c1st}{\bf{27.561}} & \SetCell{c2nd}{0.104} \\
Stump    & 4.82 & 0.776 & 26.672 & 0.209 & 4.97 & 0.783 & 26.775 & 0.196 & 5.02 & \SetCell{c1st}{\bf{0.786}} & \SetCell{c3rd}{26.866} & \SetCell{c1st}{\bf{0.195}} & 4.67 & \SetCell{c1st}{\bf{0.786}} & \SetCell{c2nd}{26.869} & \SetCell{c2nd}{0.196} & 4.93 & \SetCell{c1st}{\bf{0.786}} & \SetCell{c1st}{\bf{26.874}} & \SetCell{c1st}{\bf{0.195}} \\
Treehill    & 3.43 & \SetCell{c2nd}{0.663} & \SetCell{c3rd}{22.367} & 0.324 & 4.14 & 0.631 & 22.282 & \SetCell{c1st}{\bf{0.29}} & 3.63 & \SetCell{c2nd}{0.633} & 22.315 & \SetCell{c3rd}{0.293} & 3.8 & \SetCell{c1st}{\bf{0.635}} & \SetCell{c1st}{\bf{22.422}} & \SetCell{c2nd}{0.292} & 3.73 & \SetCell{c1st}{\bf{0.635}} & \SetCell{c2nd}{22.379} & \SetCell{c3rd}{0.293} \\
Flower    & 3.51 & 0.607 & 21.478 & 0.334 & 7.74 & 0.613 & 21.132 & 0.261 & 6.29 & \SetCell{c1st}{\bf{0.621}} & \SetCell{c2nd}{21.23} & \SetCell{c2nd}{0.259} & 6.38 & \SetCell{c1st}{\bf{0.621}} & \SetCell{c3rd}{21.21} & \SetCell{c1st}{\bf{0.258}} & 6.18 & \SetCell{c1st}{\bf{0.621}} & \SetCell{c1st}{\bf{21.251}} & \SetCell{c3rd}{0.26} \\
Room    & 1.43 & 0.919 & 31.455 & 0.218 & 3.82 & 0.926 & 31.493 & \SetCell{c1st}{\bf{0.193}} & 2.52 & \SetCell{c1st}{\bf{0.927}} & \SetCell{c3rd}{31.713} & 0.196 & 2.67 & \SetCell{c1st}{\bf{0.927}} & \SetCell{c2nd}{31.755} & \SetCell{c2nd}{0.194} & 2.55 & \SetCell{c1st}{\bf{0.927}} & \SetCell{c1st}{\bf{31.93}} & \SetCell{c3rd}{0.195} \\
Counter    & 1.06 & 0.907 & 28.88 & 0.201 & 2.29 & \SetCell{c2nd}{0.915} & \SetCell{c3rd}{29.183} & \SetCell{c1st}{\bf{0.177}} & 1.7 & \SetCell{c2nd}{0.915} & 29.178 & \SetCell{c3rd}{0.18} & 1.74 & \SetCell{c2nd}{0.915} & \SetCell{c2nd}{29.191} & \SetCell{c3rd}{0.18} & 1.74 & \SetCell{c1st}{\bf{0.916}} & \SetCell{c1st}{\bf{29.216}} & \SetCell{c2nd}{0.179} \\
Bonsai    & 1.1 & 0.948 & 32.477 & 0.207 & 2.15 & \SetCell{c1st}{\bf{0.954}} & \SetCell{c3rd}{32.947} & \SetCell{c1st}{\bf{0.158}} & 1.81
  & \SetCell{c1st}{\bf{0.954}} & 32.674 & \SetCell{c2nd}{0.161} & 1.84 & \SetCell{c1st}{\bf{0.954}} & \SetCell{c2nd}{32.95} & \SetCell{c2nd}{0.161} & 1.83 & \SetCell{c1st}{\bf{0.954}} & \SetCell{c1st}{\bf{32.952}} & \SetCell{c2nd}{0.161} \\
Kitchen    & 1.39 & 0.926 & 31.138 & 0.127 & 2.68 & \SetCell{c1st}{\bf{0.931}} & \SetCell{c2nd}{31.641} & \SetCell{c1st}{\bf{0.118}} & 2.12 & \SetCell{c1st}{\bf{0.931}} & \SetCell{c3rd}{31.463} & \SetCell{c2nd}{0.119} & 2.1 & \SetCell{c1st}{\bf{0.931}} & 31.331 & \SetCell{c2nd}{0.119} & 2.09 & \SetCell{c1st}{\bf{0.931}} & \SetCell{c1st}{\bf{31.698}} & 0.12 \\
\end{tblr}
}
\end{minipage}
\end{table*}

We evaluate our GeoTexDensifier on nine cases from Mip-NeRF 360~\cite{barron2022mip}, twenty one cases from Tanks and Temples~\cite{knapitsch2017tanks}, and two real outdoor scenes ``Wufeng Tower'' and ``Office Park'' captured by DJI Zenmuse P1 and Phantom 4 RTK respectively,
each of which is composed of multi-view digital images, corresponding camera intrinsic and extrinsic parameters, and sparse SfM map points recovered by COLMAP~\cite{schoenberger2016sfm}. Each experimental case typically contains some fully textured regions with appearance details, and some weakly textured surfaces. We implement GeoTexDensifier based on the source code of original 3DGS~\cite{kerbl20233d} and compare our implementation with original 3DGS, original 3DGS with depth prior from Depth Anything v2 (DAv2)~\cite{yang2024depth}, and three other SOTA methods Mini-Splatting-D~\cite{fang2024mini}, GeoGaussian~\cite{li2024geogaussian} and Pixel-GS~\cite{zhang2024pixel}. All the results are trained with the source code of each method conducted on a single Ubuntu18.04 server with an Nvidia GeForce RTX4090 GPU with $24$GB memory, and $500$GB RAM, with the metrics SSIM, PSNR and LPIPS given in subsection~\ref{sec:qual-quant-evaluation} to evaluate the corresponding rendering quality. We also exhibit ablation studies on key modules such as texture-aware densification, geometry-aware splitting with \textit{VDRC} and normal guidance in subsection~\ref{sec:ablation-study} to show the usefulness of each strategy on the improvements of the reconstruction quality of 3DGS models.

\subsection{Qualitative and Quantitative Evaluation}
\label{sec:qual-quant-evaluation}
Fig.~\ref{fig:teaser} already gives an indoor scene ``Church'' for comparison to SOTA works. As the comparison results of other cases shown in Fig.~\ref{fig:qual-compare} and Table~\ref{tab:quant-compare}, the original 3DGS~\cite{kerbl20233d} often lacks sufficient splats in fully textured areas due to its simple densification strategy. Pixel-GS~\cite{zhang2024pixel} and Mini-Splatting-D~\cite{fang2024mini} address this issue to some extent, by counting the contribution of each Gaussian in two different strategies to find under-densified splats for further splitting. However, Pixel-GS still struggles to recover adequate details in the subtle textures such as the carpet in Fig.~\ref{fig:teaser} and the chair in case ``Meeting Room'', and seems to perform even worse in texture detail recovery than original 3DGS in self-captured case ``Office Park'' as shown in Fig.~\ref{fig:qual-compare}. This under-densification issue might still occur because Pixel-GS simply regards the covered pixel numbers as the weights for averaging pixel gradients, which turns out to be less adaptable to complex scenes than original strategy.
On the other hand, Mini-Splatting-D employs a more explicit strategy to enforce splitting for reconstructing more texture details. Nevertheless, it suffers from over-splitting issue which further degrades reconstruction quality in areas with weak textures or insufficient training views. GeoGaussian introduces a planar regularization term to optimize the scene structure, but heavily relies on the quality of the input point cloud. Therefore, it specially uses PlanarSLAM~\cite{li2021rgb} to obtain a more evenly distributed point cloud with high-quality even in weakly textured areas, which might not be suitable for general purpose. When using sparse SfM map points as input in our experiment for fair comparison, its performance is deteriorated significantly by failing to provide sufficient splats in texture-rich areas as well. Original 3DGS with depth prior is more similar to our geometry-aware splitting strategy, and outperforms the original one without depth prior. However, due to the lack of true scale for DAv2~\cite{yang2024depth}, the source code implementation of original 3DGS aligns the DAv2 depth to the sparse SfM points to obtain a scale and offset for each frame as depth prior, whose accuracy might be affected by some unreliable SfM points or possible scale inconsistency across multiple frames to impact the final rendering quality. In comparison, our GeoTexDensifier fully combines texture-aware densification and geometry-aware splitting strategies to produce a 3DGS model consisting of cleaner, more reasonably and more accurately distributed splats without rely on truly scaled depth prior, which offers superior rendering quality compared to the other works as verified by the quantitative evaluation of metrics SSIM, PSNR and LPIPS given in Table~\ref{tab:quant-compare}. Please refer to the supplementary material for more comparison details in the videos.

\subsection{Ablation Studies}
\label{sec:ablation-study}
Fig.~\ref{fig:teaser} and Fig.~\ref{fig:texture-aware-weight} already demonstrate that our texture-aware densification effectively recovers appearance details in texture-rich areas while maintaining sparse splat distribution in low-texture regions. Fig.~\ref{fig:geo-split} illustrates how our geometry-aware splitting strategy further refines the initial placement of child Gaussians, so as to help our system to achieve 3DGS models with a more reasonable spatial distribution of splats. From more visualized comparisons and corresponding evaluation metrics presented in Fig.~\ref{fig:ablation-study} we can see that, without the geometry-aware splitting strategy, the system might still encounter issues such as floater noise due to unrestricted splat placement in low-texture areas or the regions with insufficient training views, while our GeoTexDensifier with both strategies achieves the best in both fully and weakly textured regions.

Table~\ref{tab:ablation} further provides more quantitative comparisons of ablation strategies including only geometry-aware splitting (Geo.), only texture-aware densification (Tex.), texture-aware densification with \textit{VDRC} (depth prior) (Tex. + VDRC), texture-aware densification with normal guidance (Tex. + Normal), and GeoTexDensifier (full strategies with Tex. + Geo.), on Mip-NeRF 360 dataset~\cite{barron2022mip}. We can see from the quatitative evaluation metrics with SSIM, PSNR and LPIPS in the ablation experiments that texture-aware densification strategy makes greater improvement in 3DGS quality than using only depth prior or normal prior due to better distributed splats, while texture-aware densification combined with either depth prior or normal prior further improves the reconstruction quality with more accurate optimization of splat positions in textureless areas. Finally, full strategies combining both depth prior and normal prior with texture-aware densification can make the best performance in the final rendering quality.

\subsection{Limitations}
\label{sec:limitations}
Our geometry-aware splitting strategy effectively resolves weakly-textured issues by filtering erroneous Gaussian splats via \textit{VDRC} checking. However, this strategy might introduce a limitation in object-centric scenarios captured with distant backgrounds such as the two outdoor cases of ``Horse'' and ``Truck'' from Tanks and Temples dataset~\cite{knapitsch2017tanks} shown in Fig.~\ref{fig:failure}. Initial SfM points for such distant regions are usually sparse or missing. When Gaussian splats in foreground regions undergo splitting, the large relative depth ratio changes on foreground/background boundary regions typically fail to pass the \textit{VDRC} checking, which consequently prevent these splats to be split to the distant regions, leading to ``under-reconstruction'' regions with missing Gaussians. These incomplete regions possibly cause blurs or ghosting artifacts in the distant backgrounds from certain viewpoints, as exemplified in the magnified regions of the representative views of the two cases in Fig.~\ref{fig:failure}.

\subsection{VR Application}
We can also import the 3DGS model generated by our GeoTexDensifier into an open-sourced ``Gaussian Splatting VR Viewer Unity Native Plugin''~\footnote{https://github.com/clarte53/GaussianSplattingVRViewerUnity} to achieve photorealistic immersive visualization and real-time interactive navigation of the reconstructed scene on an Oculus Quest II standalone VR headset equipped with a single Ubuntu18.04 desktop PC with Intel Core i7-8700 CPU @ 3.2GHz, an Nvidia GeForce RTX4090 GPU, and $48$GB RAM, as illustrated in Fig. \ref{fig:VR-application}. Note that the example of the ``Wufeng Tower'' can be immersively navigated by 3DGS representation with highly restored geometry and texture details. Please refer to the supplementary video for the complete demo.

\section{Conclusion and Future Work}
We introduce GeoTexDensifier, a novel framework designed to more effectively enhance the densification of 3DGS, which begins with a texture-aware densification strategy that leverages texture gradients from training images to ensure the splats to be adequately populated in highly-textured areas while maintaining sparsity in regions with low textures, followed by employing a geometry-aware splitting module to guide the spatial distribution of splats to be better aligned with the actual surface of the scene by validating the initially split splat positions according to the relative depth priors. As a result, a high-quality 3DGS model is produced with more reasonably distributed splats and less noise compared to SOTA methods, leading to more photorealistic NVS effects.

Our limitation in object-centric scenes with distant backgrounds discussed in section~\ref{sec:limitations} can be better solved by combining a more reliable dense point cloud initialization instead of sparse SfM points to further avoid ``under-reconstruction'' situation. Besides, like other SOTA methods, our approach introduces more splats to reconstruct finer details in textured areas, which results in higher GPU memory consumption than original 3DGS when handling large-scale scenes. Our \textit{VDRC} checking and normal guidance operations also increase the training time complexity. We plan to extend our method to support urban-scale reconstruction and navigation in a more space/time-efficient way by integrating techniques such as chunk-based reconstruction, LOD organization and dynamically scheduled rendering. Additionally, the potential possibility of combining our approach with generative diffusion models will be explored as a future work to further enhance reconstruction completion and quality in insufficiently observed or invisible regions.

\section*{Acknowledgments}
The authors wish to thank Zhuang Zhang, Chongshan Sheng, Fei Jiao, Xia Sun, Jinyi Liao, Renzhi Wang, Yuqing Xie, Bingze Li and Hongliang Sun for their kind helps in the development and the experiments of the proposed GeoTexDensifier system. This work was partially supported by Key R\&D Program of Zhejiang Province (No. 2023C01039), and NSF of China (No. 62425209).

\bibliographystyle{IEEEtran}

\newpage

 
\vspace{11pt}

\begin{IEEEbiography}[{\includegraphics[width=1in,height=1.25in,clip,keepaspectratio]{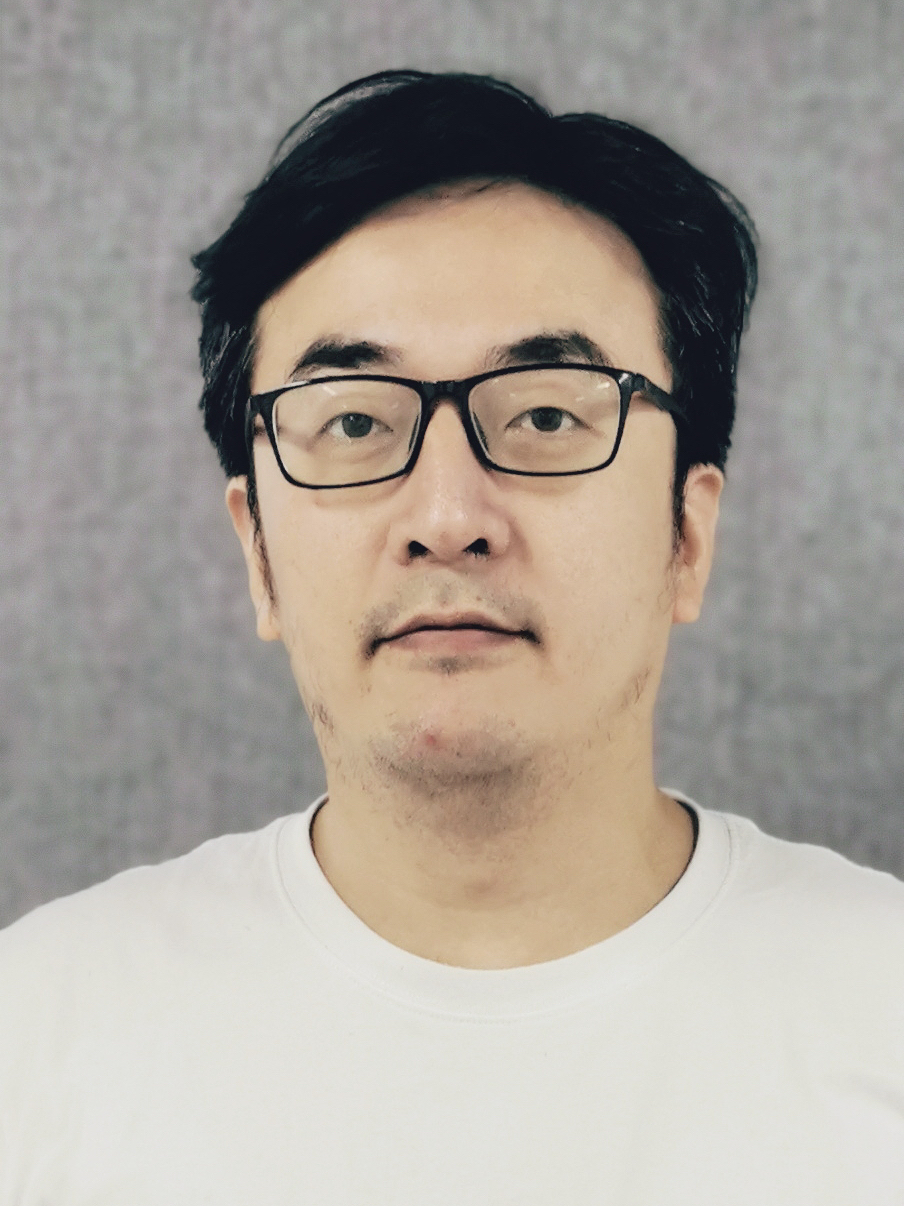}}]{Hanqing Jiang}
received his Ph.D. degree from Zhejiang University, after which he was a postdoctoral researcher in the State Key Lab of CAD\&CG, Zhejiang University. He is currently a Research Director in SenseTime Group Ltd, China. He received the best paper award of ISMAR 2020 and the best journal paper nominee of ISMAR 2021. His research interests focus on computer vision, including video enhancement, multi-view stereo, 3D reconstruction, neural radiance fields, and augmented reality.
\end{IEEEbiography}

\vspace{11pt}

\begin{IEEEbiography}[{\includegraphics[width=1in,height=1.25in,clip,keepaspectratio]{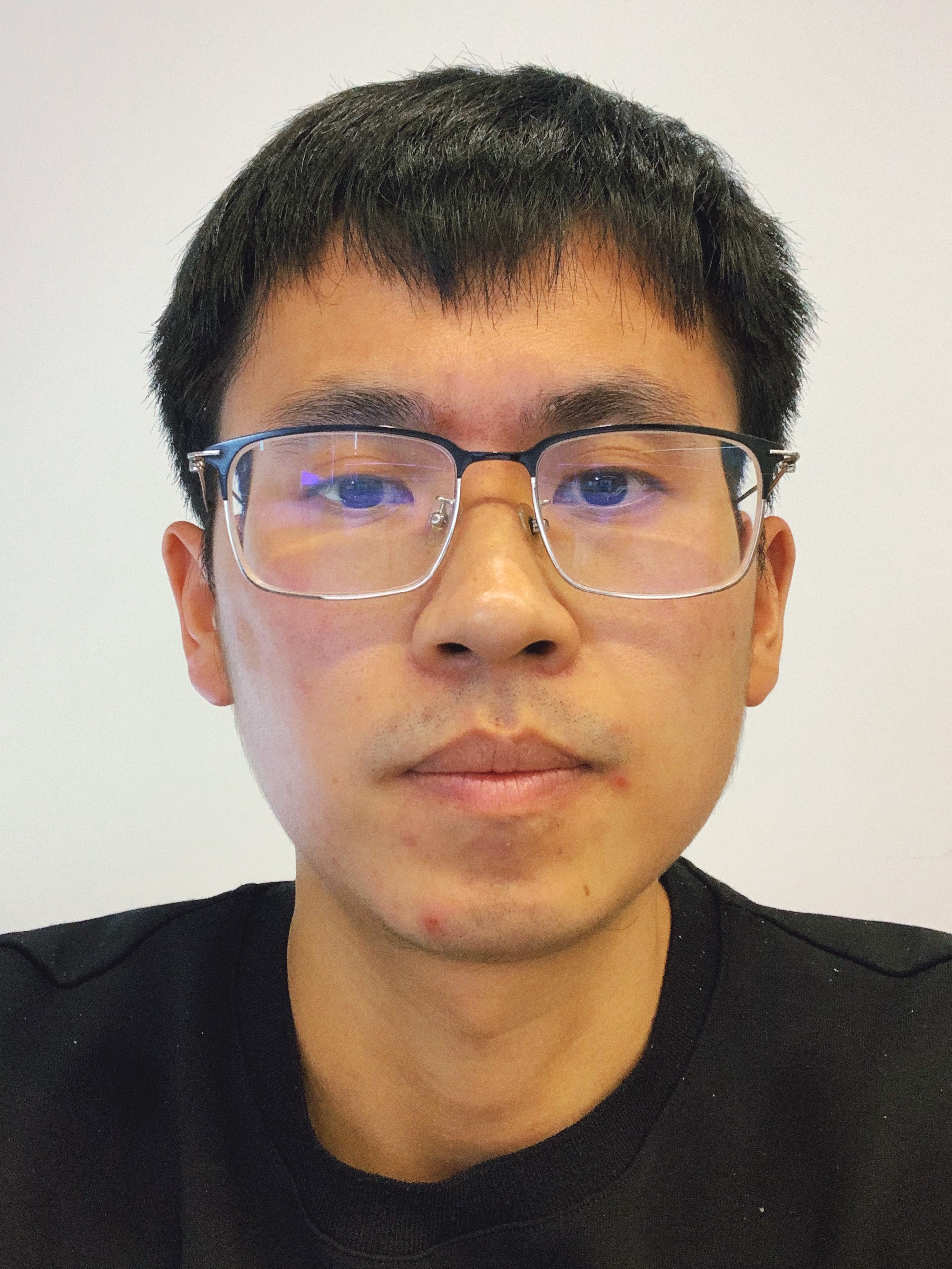}}]{Xiaojun Xiang}
received his master degree from the State Key Lab of CAD\&CG, Zhejiang University. He is currently a Senior Researcher in SenseTime Group Ltd, China. He received the best journal paper nominee of ISMAR 2021. His research interests lie in the field of 3D computer vision, including multi-view stereo, vectorized modeling, and implicit neural reconstruction.
\end{IEEEbiography}

\vspace{11pt}

\begin{IEEEbiography}[{\includegraphics[width=1in,height=1.25in,clip,keepaspectratio]{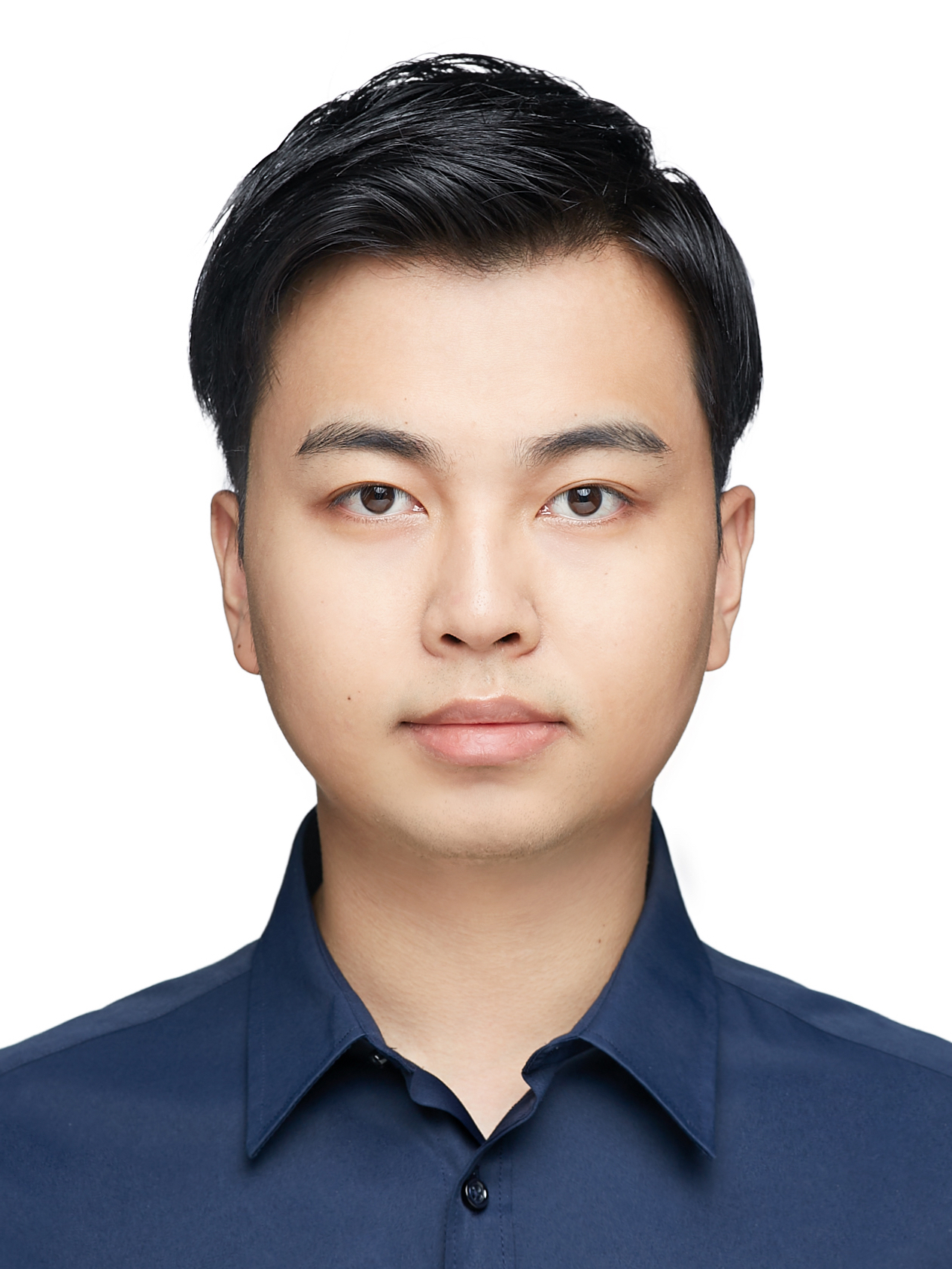}}]{Han Sun}
is currently a researcher in SenseTime Group Ltd, China. He received his master degree from University of Science and Technology of China in 2018. His research interests include structure from motion, multi-view stereo, and 3D reconstruction.
\end{IEEEbiography}

\vspace{11pt}

\begin{IEEEbiography}[{\includegraphics[width=1in,height=1.25in,clip,keepaspectratio]{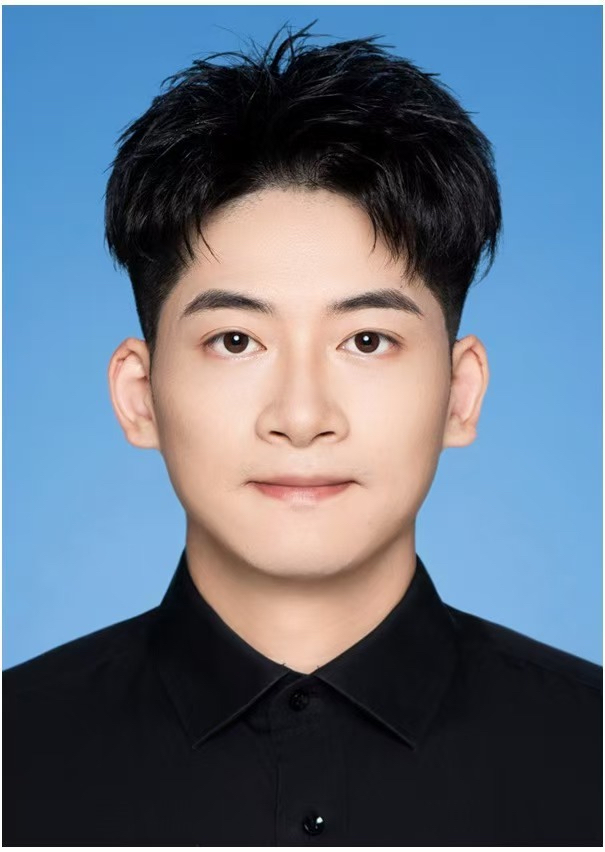}}]{Hongjie Li}
received the M.S. and Ph.D degrees in photogrammetry and remote sensing from the State Key Laboratory of Information Engineering in Surveying, Mapping and Remote Sensing (LIESMARS), Wuhan University, Wuhan, China, in 2020 and 2024, respectively. He is currently a Researcher in SenseTime Group Ltd, China. His research interests include multi-view stereo, 3D reconstruction and 3D generation.
\end{IEEEbiography}

\vspace{11pt}

\begin{IEEEbiography}[{\includegraphics[width=1in,height=1.25in,clip,keepaspectratio]{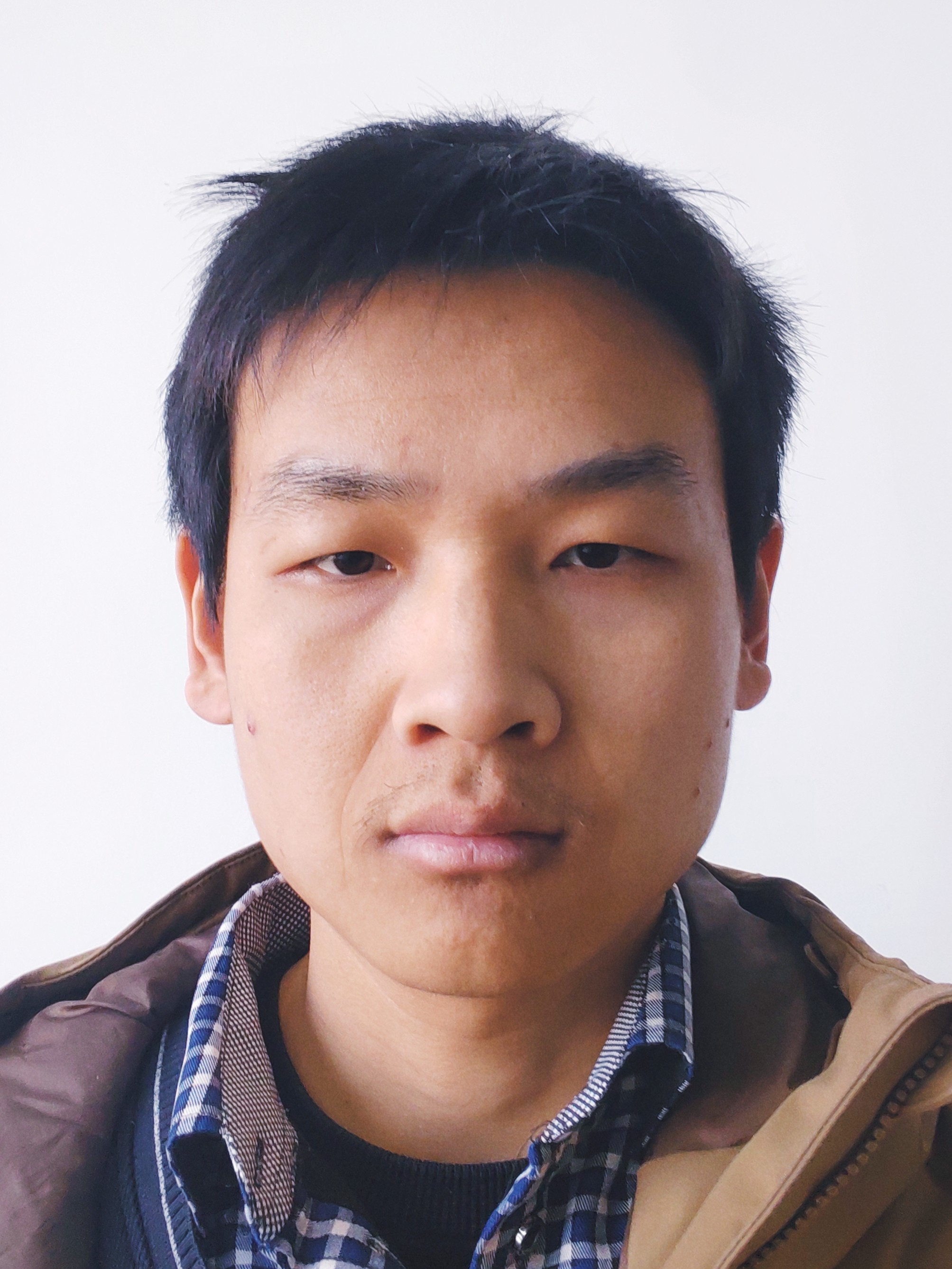}}]{Liyang Zhou}
is currently a Senior Algorithm Engineer in SenseTime Group Ltd, China. He received his master degree from Zhejiang University in 2016, and the best paper award of ISMAR 2020. His research interests include structure from motion, multi-view stereo, and 3D reconstruction.
\end{IEEEbiography}

\vspace{11pt}

\begin{IEEEbiography}[{\includegraphics[width=1in,height=1.25in,clip,keepaspectratio]{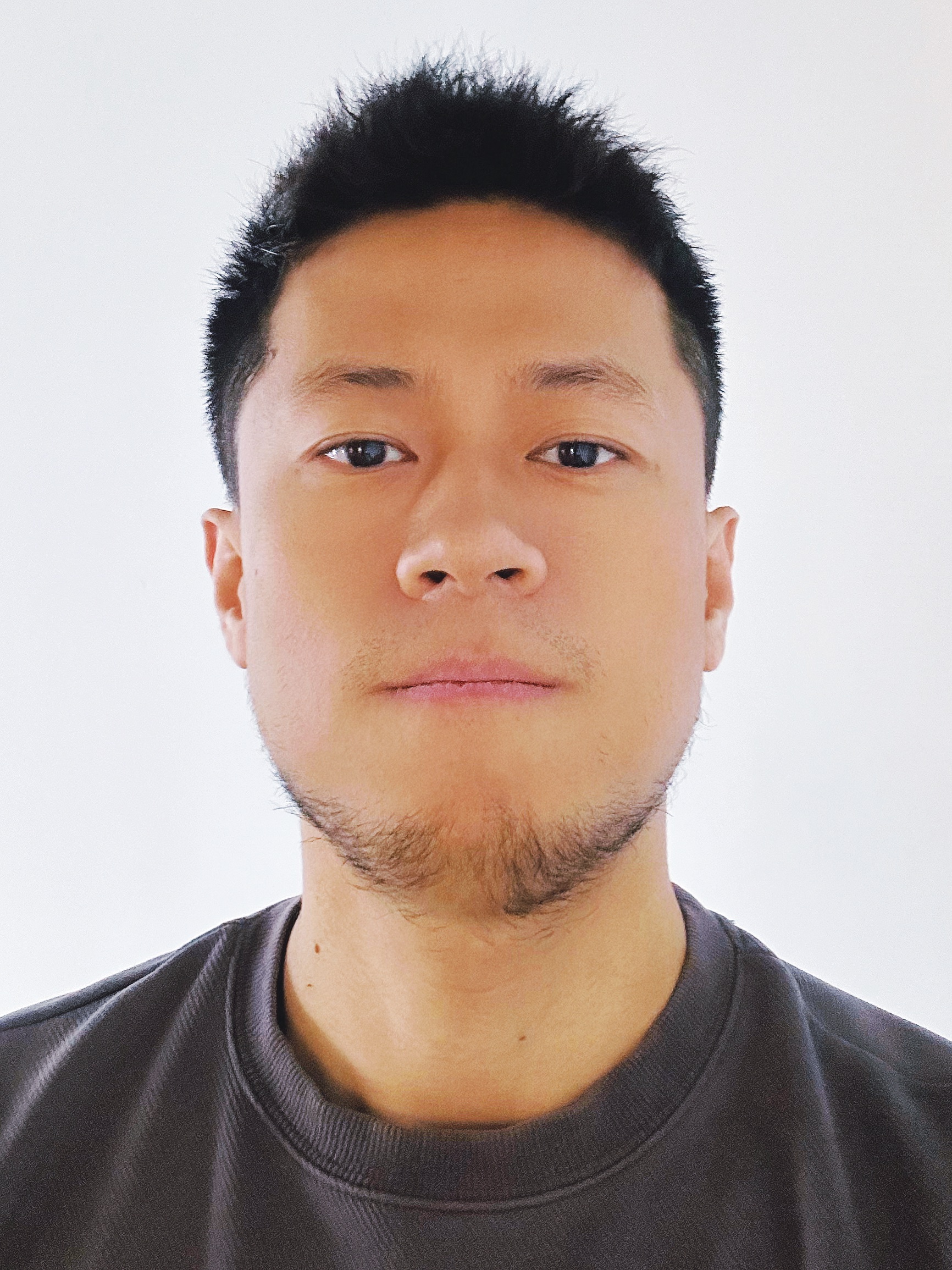}}]{Xiaoyu Zhang}
is a researcher at SenseTime Group Ltd, China. He received his master's degree from Northwestern Polytechnical University in 2019. His research interests include novel view synthesis and 3D reconstruction.
\end{IEEEbiography}

\vspace{11pt}



\begin{IEEEbiography}[{\includegraphics[width=1in,height=1.25in,clip,keepaspectratio]{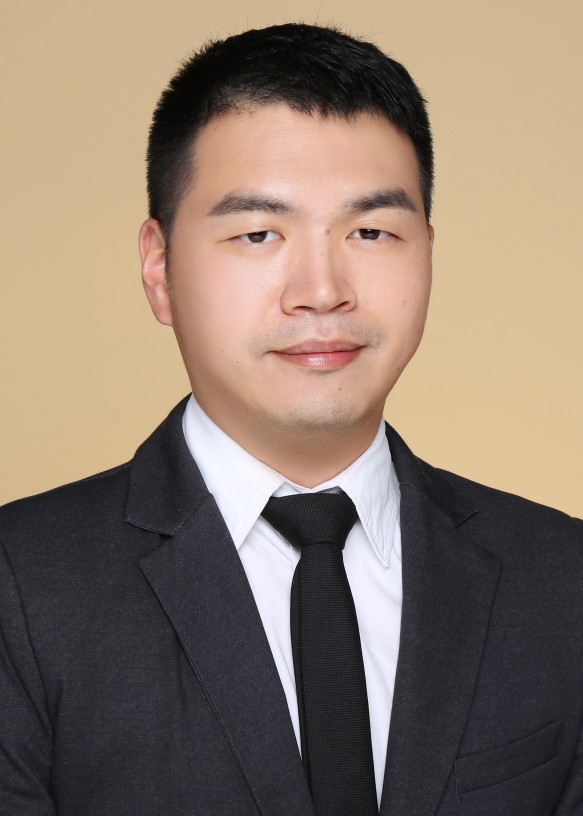}}]{Guofeng Zhang}
is now a professor at State Key Lab of CAD\&CG, Zhejiang University. He received his BS and Ph.D. degrees in Computer Science from Zhejiang University, in 2003 and 2009, respectively. He received the National Excellent Doctoral Dissertation Award, the Excellent Doctoral Dissertation Award of China Computer Federation and the best paper award of ISMAR 2020. His research interests include structure-from-motion, SLAM, 3D reconstruction, augmented reality, spatial intelligence.
\end{IEEEbiography}


\vfill

\end{document}